\documentclass[]{jfm}
\pdfoutput=1
\usepackage{graphicx}
\usepackage{newtxtext}
\usepackage{newtxmath}
\usepackage{natbib}
\usepackage{hyperref}
\hypersetup{
    colorlinks = true,
    urlcolor   = blue,
    citecolor  = black,
}

\newcommand{\RomanNumeralCaps}[1]
\linenumbers

\usepackage[ruled,boxed]{algorithm2e}
\usepackage{makecell}
\usepackage{graphicx}
\usepackage{bm,booktabs,bigstrut}
\usepackage{overpic}
\usepackage{multirow} 
\usepackage{rotating}
\newcommand{\rot}[2]{\begin{turn}{#1}#2\end{turn}}

\usepackage{tabularx}
\newcolumntype{P}[1]{>{\centering \arraybackslash}p{#1}}
\newcolumntype{L}{X}
\newcolumntype{C}{>{\centering \arraybackslash}X}
\newcolumntype{R}{>{\raggedright \arraybackslash}X}

\newcommand{\layout}[2]{$\mathcal{L}_{\uppercase\expandafter{\romannumeral#1}}^{#2}$}
\newcommand{\ie}{\textit{i}.\textit{e}., }
\newcommand{\fig}[1]{(\textit{#1})}
\newcommand{\dd}[1]{$\mathrm{d}#1$}

\usepackage[x11names]{xcolor}

\title{Dynamic feature-based deep reinforcement learning for flow control of circular cylinder with sparse surface pressure sensing}

\author{
    Qiulei~Wang\aff{1},
    Lei~Yan\aff{1} \corresp{\email{180410212@stu.hit.edu.cn}},
    Gang~Hu\aff{1,2,3} \corresp{\email{hugang@hit.edu.cn}},
    Wenli~Chen\aff{3,4},
    Jean~Rabault\aff{5},
    \and Bernd~R.~Noack\aff{6}
}

\affiliation{
    \aff{1}School of Civil and Environmental Engineering, Harbin Institute of Technology, Shenzhen 518055, China
    \aff{2}Guangdong Provincial Key Laboratory of Intelligent and Resilient Structures for Civil Engineering, Harbin Institute of Technology, Shenzhen, 518055, China
    \aff{3}Guangdong-Hong Kong-Macao Joint Laboratory for Data-Driven Fluid Mechanics and Engineering Applications, Harbin Institute of Technology, Shenzhen 518055, China
    \aff{4}Key Laboratory of Smart Prevention and Mitigation of Civil Engineering Disasters, the Ministry of Industry and Information Technology, Harbin Institute of Technology, Harbin, 150090, China
    \aff{5}Independent Researcher, Oslo 0376, Norway
    \aff{6}School of Mechanical Engineering and Automation, Harbin Institute of Technology, Shenzhen 518055, China
}

\begin{document}
\maketitle

\begin{abstract}
This study proposes a self-learning algorithm for closed-loop cylinder wake control targeting lower drag and lower lift fluctuations with the additional challenge of sparse sensor information, taking deep reinforcement learning as the starting point. DRL performance is significantly improved by lifting the sensor signals to dynamic features (DF), which predict future flow states. The resulting dynamic feature-based DRL (DF-DRL) automatically learns a feedback control in the plant without a dynamic model. Results show that the drag coefficient of the DF-DRL model is 25\% less than the vanilla model based on direct sensor feedback. More importantly, using only one surface pressure sensor, DF-DRL can reduce the drag coefficient to a state-of-the-art performance of about 8\% at \Rey = 100 and significantly mitigate lift coefficient fluctuations. Hence, DF-DRL allows the deployment of sparse sensing of the flow without degrading the control performance. This method also exhibits strong robustness in flow control under more complex flow scenarios, reducing the drag coefficient by 32.2\% and 46.55\% at \Rey =500 and 1000, respectively. Additionally, the drag coefficient decreases by 28.6\% in a three-dimensional turbulent flow at \Rey =10000. Since surface pressure information is more straightforward to measure in realistic scenarios than flow velocity information, this study provides a valuable reference for experimentally designing the active flow control of a circular cylinder based on wall pressure signals, which is an essential step toward further developing intelligent control in realistic multi-input multi-output (MIMO) system.
\end{abstract}



\section{Introduction}\label{sec:intro}
Flow control has been a popular research area of great academic and industrial interest, which can be divided into passive flow control and active flow control on the basis of whether external energy input is necessary. Passive control has the advantages of requiring no energy, being easy to set up, and having low cost, but if the actual situation of the flow field differs from that expected, the control is often difficult to achieve the best effect. Active control can be divided into open-loop control and closed-loop control according to whether it is necessary to obtain feedback information from the flow field and adjust the flux of the actuator \citep{duriez_machine_2017}. It has been found that compared with open-loop control, closed-loop active control has a robust adaptive ability, which can give full play to the potential of the actuator with a small amount of energy input. For example, \citet{korkischko2012suppression} has presented a sequence of experiments on the flow around a cylinder, which included MSBC (Moving Surfaces Boundary Control). This method involved injecting momentum into the boundary layer of the cylinder by using two rotating cylinder modules, thereby delaying separation and preventing vortices. Nevertheless, the complexity of the nonlinear Navier-Stokes equations leads to a flow field with high dimension and multimodal characteristics, thus making it challenging to devise effective real-time closed-loop active flow control procedures.

In recent years, machine learning has made significant advances, and active flow control is becoming more effective and intelligent \citep{Brunton2015amr}. One of the earliest machine learning techniques applied in this field was genetic programming (GP). GP uses a population of computer programs as potential solutions to a problem. The programs evolve using genetic operators like mutation and crossover, and the fittest ones produce the next generation of solutions. \citet{gautier2015closed} applied GP to search explicit control laws for reducing the recirculation zone behind a backward-facing step. \citet{zhou2020artificial} applied the linear GP to control the dynamics of a turbulent jet and discovered novel wake patterns. \citet{ren2019active} adopted GP-identified control laws to successfully suppress vortex-induced vibrations in a numerical simulation environment. \citet{pino2023comparative} demonstrated that many techniques from the ML family could be applied to AFC tasks, from GP to Bayesian Optimization (BO) \citep{ji2022active,blanchardBayesianOptimizationActive2021}, Lipschitz Global Optimization (LIPO) \citep{pinter1995global}, and Reinforcement Learning (RL) \citep{mnihHumanlevelControlDeep2015a,rabaultDeepReinforcementLearning2020,wang2022accelerating}, and that these methods have trade-offs relatively to each other.

Artificial neural networks (ANNs) can also be trained to learn complex patterns and relationships in fluid dynamics data and to generate control strategies that optimize fluid manipulation, which can be used for various tasks, including predicting fluid flow patterns, controlling robotic arms that manipulate fluids, and optimizing the design of microfluidic devices. \citet{lee_application_1997} applied an adaptive controller based on a neural network for turbulent channel flow, demonstrating a simple control scheme that reduced skin friction by up to 20\% and produced an optimum wall blowing and suction proportional to a local sum of wall-shear stress. 

The rapid development of deep reinforcement learning (DRL), which is effective at interacting with complex nonlinear environments, has brought new ideas to the above flow control problems. Previous studies have shown that deep reinforcement learning can effectively acquire control strategies in high-dimensional, non-linear, and other complex environments. Suppose that deep reinforcement learning is employed to interact with a flow control environment. In such a scenario, it is essential for the closed-loop flow control method to establish the control law based on the learned strategy after continuous trial and error and adjustment of the optimization strategy. 

\citet{Rabault2019Artificial} made a groundbreaking contribution by introducing DRL to active flow control for the first time by applying deep reinforcement learning to blunt body drag reduction at $Re = 100$ and successfully demonstrated a closed-loop active control strategy that could achieve stable drag reduction of about 8$\%$ by using proximal strategy optimization (PPO) method. This study uses the velocity measured by 151 sensors around the cylinder and in the downstream flow field (each sensor collects both the flow lateral velocity) as the feedback signal. To investigate higher Reynolds numbers, \citet{ren2021applying} applied the Lattice-Boltzmann Method (LBM) to establish a CFD environment with weak turbulence conditions, and a Reynolds number of up to 1000 was effectively controlled. The results show that the DRL agent could find an effective feedback law and achieve a drag reduction of more than 30$\%$. Applications in even more chaotic conditions, corresponding to a 2D cylinder at a Reynolds number $Re = 2000$, have recently been presented by \citet{varela2022deep}, highlighting that DRL-based controllers can learn drastically different control laws as the underlying dominating physics are changed. In another study, \citet{tang2020robust} placed four synthetic jets on the lower and upper sides symmetrically for active flow control of the cylinder.

Apart from the jet actuator, two small rotating cylinders were placed obliquely behind the main cylinder at a Reynolds number of 240 in \citet{xu2020active}'s study. The rotational speed of the small cylinders was controlled by a DRL agent. This experimental setup aimed to investigate the potential of wake stabilization using DRL-controlled rotating control cylinders. The study findings were later confirmed by \citet{fanReinforcementLearningBluff2020}, who experimentally verified the effectiveness and feasibility of this approach.

In addition to its application in the field of AFC tasks, researchers have also aimed to utilize the DRL approach to achieve other objectives. These objectives include reducing the energy expenditure of the follower \citep{novati2017synchronisation,verma2018efficient}, mitigating vortex-induced vibration \citep{ren2019active,ren2021bluff,mei2021active,zheng2021active}, shape optimization \citep{garnier2021review,viquerat2021direct,li2021knowledge}, or the control of turbulent channel flows \citep{guastoni2023deep}. As the field of DRL applications for fluid mechanics is evolving fast, we refer the reader curious for more details to any of the recent reviews on the topic, \textit{i.e.} \citet{garnier2021review,vignon2023recent}.

Most of the aforementioned studies have collected state information using a large number of velocity sensors in the wake region, which poses significant challenges for practical structural flow fields. For instance, in the case of vehicles and high-rise buildings, it would be more convenient and easier to maintain and deploy surface pressure sensors. However, compared to the state in the wake region, the pressure on the surface of the structure may have insufficient characteristic information, making it difficult for the DRL agent to estimate the state of the entire flow field. This will result in typical reinforcement learning methods being unable to learn effective control strategies. Based on this fact, we introduce the dynamic feature (DF) lifting approach to deep reinforcement learning and propose the DF-DRL method. In the case of flow around the cylinder, this method can significantly enhance the convergence performance of the DRL algorithm, enabling it to achieve a drag reduction effect that is almost consistent with the benchmark (147 velocity sensors deployed in the wake region) with a reduction of 99.3\% of the sensor quantity.

In the present study, we utilized the DRLinFluids package \citep{wang2022drlinfluids} to train a DRL agent and execute interactions. The package leverages Tensorforce \citep{tensorforce} and Tianshou \citep{tianshou} packages to provide DRL algorithm libraries, and OpenFOAM \citep{jasak2007openfoam} as the CFD interaction environment. Firstly, we compare the performance of vanilla DRL and DF-DRL-based plants to a benchmark case study of flow around a circular cylinder with a Reynolds number of 100. Subsequently, we varied the number of pressure surface sensors to validate the effectiveness of the proposed method. Finally, we trained a DF-DRL agent with a single surface pressure sensor and deployed it to more complex flows at higher Reynolds numbers to illustrate the robustness of the approach.

\section{Active flow control system with DRL-based jet actuators} \label{sec:Methodology}
The present section is partitioned into two components: (1) an illustration of the DRL algorithm, especially for the Soft Actor-Critic (SAC) method, which will be used as the DRL part in the whole study; (2) a detailed introduction of the dynamic feature-based DRL framework, including the dynamic feature lifting and the coupling with the flow simulation.

\subsection{Deep reinforcement learning} \label{sec:DRL}
Deep reinforcement learning (DRL) is a powerful method of optimal control based on a parameterized policy, commonly called an agent, that learns through trial and error. In the context of computational fluid dynamics (CFD), the environment can be modeled as the flow over a circular cylinder. During the optimization procedure, the DRL agent interacts with this environment to generate experiences according to the current policy. These experiences are then cached in a buffer and used by the training algorithm to improve the policy. This iterative process is repeated until the agent can yield a control strategy that satisfies the desired performance criteria. Thus, DRL has the potential to revolutionize the field of fluid mechanics by enabling the discovery of previously unknown control strategies that can enhance the performance of fluid systems.

There are several types of DRL algorithms \citep{francoislavetIntroductionDeepReinforcement2018}. One of the most popular types of DRL algorithms is Q-learning \citep{mnihPlayingAtariDeep2013,bellemareDistributionalPerspectiveReinforcement2017,andrychowiczHindsightExperienceReplay2018}, which uses a neural network to approximate the optimal action-value function, and updates the network's weights using the Bellman equation to minimize the difference between the predicted and actual reward. Another type of algorithm is policy gradient methods \citep{mnihAsynchronousMethodsDeep2016,Schulman2017Proximal,schulmanTrustRegionPolicy2017}, which directly optimize the agent's policy to maximize the expected reward and often use techniques like Monte Carlo sampling or trust region optimization. Actor-critic methods \citep{lillicrapContinuousControlDeep2019a,fujimotoAddressingFunctionApproximation2018,haarnoja2018soft} combine the advantages of both Q-learning and policy gradient methods by simultaneously learning a value function and a policy. Another type of DRL algorithm is model-based reinforcement learning \citep{haWorldModels2018,weberImaginationAugmentedAgentsDeep2018,silverMasteringChessShogi2017}, which involves learning a model of the environment dynamics and using it to plan actions. Model-based algorithms can be more sample-efficient than model-free algorithms like Q-learning but require additional computational resources to learn and maintain the model. Referring to our previous work \citep{wang2022drlinfluids}, the SAC algorithm is a feasible choice selected in the following study.

The Soft Actor-Critic (SAC) method \citep{haarnoja2018soft} is an actor-critic off-policy DRL algorithm that learns by leveraging a maximum entropy reinforcement learning algorithm. The agent aims to maximize entropy and prospective reward and reach the desired value while acting as randomly as possible. Since it is an off-policy algorithm, training can be performed efficiently with limited samples. The optimal policy can be formulated as
\begin{gather}
\pi^{*}=\arg \max_{\pi(\theta)} \mathbb{E}_{\left(s_{t}, a_{t}\right) \sim \rho_{\pi}}[\sum_{t} \underbrace{R\left(s_{t}, a_{t}\right)}_{\text {Reward}}+\alpha \underbrace{H\left(\pi_\theta \left(\cdot \mid s_{t}\right)\right)}_{\text{Entropy}}], \\
H\left(\pi_\theta \left(\cdot \mid s_{t}\right)\right)=-\sum_{i=1}^n p\left(\pi_\theta \left(\cdot \mid s_{t}\right)\right) \log p\left(\pi_\theta \left(\cdot \mid s_{t}\right)\right),
\label{con:rewardinDRL}
\end{gather}
where $R(s_t, a_t)$ represents the reward for taking action $a_t$ in state $s_t$, $H(\pi_{\theta}(\cdot|s_t))$ is the entropy of the policy $\pi_{\theta}$ at state $s_t$, and $\alpha$ is the temperature coefficient. The smaller $\alpha$ is, the more uniform the distribution of the output action becomes, and the maximum entropy reinforcement learning degrades to standard reinforcement learning (i.e., $\alpha \rightarrow 0$). The objective is to find the optimal policy $\pi_{\theta}$ that maximizes the expected reward while maximizing entropy.

The core idea behind the maximum entropy approach is to randomize the policy by distributing the probability of each action output widely rather than concentrating on a single action. This approach enables the neural network to explore all possible optimal paths and avoid losing the essence of maximum entropy to a single action or trajectory. The resulting benefits include (1) Learning policies that can serve as initializations for more complex tasks, as the policy learns multiple ways to solve a given task, making it more conducive to learning new tasks. (2) Strengthening the ability to explore makes identifying better patterns easier under multimodal reward conditions. (3) Enhancing the robustness and generalization ability of the approach since the optimal possibilities are explored in different ways, making it easier to adjust in the presence of interference.

The entropy term in the SAC algorithm affects the policy's exploration in two important ways. First, the entropy term encourages the policy to take more exploratory actions by adding a penalty to the objective function for actions with low probability under the current policy. This penalty is proportional to the negative entropy of the policy, which measures the degree of randomness or uncertainty in the actions selected by the policy. Minimizing this penalty incentivizes the policy to explore more widely and try out new actions that may lead to higher rewards. Second, the entropy term also helps prevent the policy from becoming too deterministic, which can limit its ability to adapt to changes in the environment or learn new behaviors. By adding an entropy term to the objective function, SAC encourages the policy to maintain a balance between exploration and exploitation rather than becoming overly focused on a single optimal action. This can be particularly important in environments with multiple suboptimal solutions or where the optimal solution may change over time. In summary, the entropy term in SAC encourages exploration by penalizing the policy for taking low-probability actions. It helps prevent the policy from becoming too deterministic by promoting a balance between exploration and exploitation. This can lead to better performance and more robust learning in complex, dynamic environments.

The SAC algorithm used in this study employs a maximum entropy target as its optimization objective, which has been shown to enhance the algorithm's exploration properties and robustness. The ability to effectively explore is achieved by maximizing information entropy, which promotes a uniform distribution of the probability of each action output rather than focusing on a single action. For instance, the uniform strategy is a high-entropy strategy. On the other hand, the robustness of the algorithm is reflected in its ability to generate alternative action outputs when faced with environmental noise. In contrast, a previous greedy strategy may lead to the agent's inefficacy due to the certainty of its actions. The SAC algorithm ensures that every action has a varying probability rather than being either high or low. Therefore, the agent can still produce alternative action outputs without failure when the environment encounters noise. From these perspectives, the SAC algorithm is highly suitable for the present study, which involves the application of active flow control (AFC) using surface pressure temporal series. For a more detailed description of the SAC algorithm, please refer to \citet{haarnoja2018soft}.

\begin{figure}
\centerline{\includegraphics[width=0.9\textwidth]{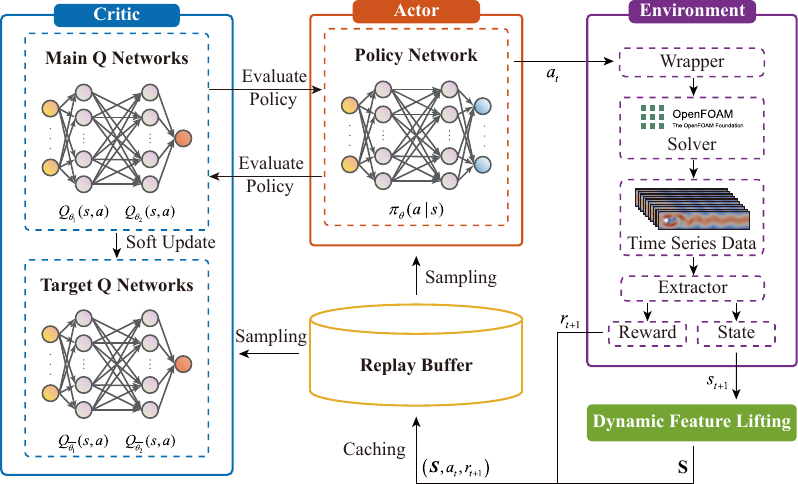}}
\caption{Schematic of the DF-DRL (SAC) framework used in the present study. The term \textit{wrapper} refers to the process of encapsulating actions from the agent and sending them to the OpenFOAM solver. In contrast, \textit{extractor} refers to the process of parsing CFD results and providing feedback to the agent. This framework is derived from the DRLinFluids package \citep{wang2022drlinfluids}.}
\label{fig:DRL_step}
\end{figure}

The present work employs a closed-loop control framework for the AFC task described in \S \ref{sec:CFD}, as depicted in Fig. \ref{fig:DRL_step}. The framework comprises two main components: the environment and the DRL agent (critic and actor in the case of the SAC algorithm used). The flow around a circular cylinder is simulated by OpenFOAM as has been described in \S \ref{sec:CFD}. The flow velocity or pressure measured by specific sensors is collected as the state provided to the agent. Following the setup of sensors in \citet{Rabault2019Artificial}'s study, 147 sensors can capture sufficient flow information for control policy learning, which is adapted as a baseline. The DRLinFluids package \citep{wang2022drlinfluids} is used to accomplish the interaction between the DRL agent and the CFD simulation, and Tianshou \citep{tianshou} is employed as the DRL algorithm backend. The DRL policy network consists of two dense layers, each with 512 fully connected neurons. The input layer receives data from pressure sensors, and the output layer gives the jet velocity. The time interval between each step is set to $7.5\%$ of the vortex-shedding period of the cylinder without actuators. The SAC agent interacts with and updates the ANN parameters every 50 steps. The process is repeated three times with the same hyperparameters to ensure the stability and validity of the training. To save training time and provide a consistent start point, a vanilla case without control is simulated in advance until it reaches stable status. Then, the state of the flow field is stored and utilized as the initialization for the following DRL training stage.

To avoid non-physical abrupt changes in pressure and velocity resulting from the use of incompressible CFD algorithms, a continuous-time approach is adopted for the control mechanism. The control for each jet is determined at every time step during the simulation and smoothed to obtain a continuous control signal over time. An appropriate interpolation method is crucial to serialize this system's received time-discretized control signal. Hence, we smooth the jet actuation to ensure a continuous change in the control signal without excessive lift fluctuations due to sudden changes in jet velocity. Based on the interpolation functions demonstrated by \citet{tang2020robust}, the control action is set to change as follows:
\begin{equation}
    V_{\Gamma_i(t)} = V_{\Gamma_i(t-1)}+ \alpha[a - V_{\Gamma_i(t-1)}],\quad i=1, 2,
\label{eq:action_process}
\end{equation}
where $\alpha = 0.1$ is a numerical parameter determined by trial and error, $V_{\Gamma_i(t)}$ and $V_{\Gamma_i(t-1)}$ is the jet flow velocity used at the non-dimensional times $t$ and $t-1$ respectively, and $a$ is one jet flow velocity in an agent step, \textit{i.e.} the action generated by the DRL agent.

Flow control of circular cylinders is a highly popular topic in both academic and industrial sectors. This research aims to reduce or eliminate drag and lift forces using advanced reinforcement learning techniques. This objective can be achieved by setting an appropriate reward function. A reward function that combines drag and lift coefficients is proposed to achieve the optimization goal, which is designated as follows:
\begin{equation}
    r_t = (C_D)_{Baseline} - \langle C_D^{t} \rangle_T - 0.1*|\langle C_L^{t} \rangle _T|, \label{eq:reward_function}
\end{equation}
where $(C_D)_{Baseline}$ is the mean drag coefficient of the circular cylinder without flow control, $C_D^{t}$ and $C_L^{t}$ means temporal drag and lift coefficient at the time $t$ respectively, and $\langle \cdot \rangle _T$ indicates the sliding average back in time over a duration corresponding to one jet flow control period $T$ with active flow control. It consists of three parts: (1) the mean drag coefficient of the non-controlled flow around a cylinder, serving as a baseline for the reward function, which helps to shift the overall mathematical expectation of the function closer to zero. This prevents the probability of sampling a potentially optimal action from decreasing as the gradient ascent progresses in the action space \citep{weaverOptimalRewardBaseline2013}; (2) the moving average drag coefficient with two DRL-based actuators that is the main component of the overall reward function. It aims to indicate to the DRL agent that its primary objective is to reduce drag force (coefficient); and (3) the absolute value of a scaled moving average lift coefficient as a penalty term. In the absence of this penalization, the policy network within the SAC algorithm has the potential to manipulate the flow pattern in order to achieve a greater reduction in drag, reaching up to approximately 18\% drag reduction \citep{rabault2019accelerating} in some cases. However, this comes at a significant trade-off as it also results in a substantial increase in induced lift, which is detrimental in the majority of practical applications.

\subsection{Active flow control with dynamic feature-based DRL enhancement} \label{sec:Preprocessing}

CFD numerical simulations allow the collection of space-time-resolved data within the considered computational domain. However, in the real world, obtaining a comprehensive view of the flow field is often difficult, which means only a limited number of time-resolved sensor measurements $\bm{s}$ are accessible. This study aims to demonstrate how recent advancements in system identification and machine learning can be utilized to construct reduced-order models directly from these sparse sensor measurements. To achieve this, we simulate experimental conditions using direct numerical simulations and focus on a single sensor measurement represented by
\begin{equation}
    s\left( t\right) :=p\left( t;\mathcal{L}\right),
\end{equation}
where $p$ is the surface pressure. The measurement vector $s$ can generally comprise various measurements such as the lift and drag coefficients, pressure measurements on a cylinder, or velocity field measurements at specific locations, e.g., wake region. However, for the scope of this study, the pressure alone is deemed adequate to characterize the flow according to the results shown in \S \ref{sec:DF-DRL_Results}.

Given the sensor measurements $\bm{s}$, our objective is to develop a practical flow state estimation that enables a DRL agent to obtain efficient information based on it. However, raw signals may not be ideal for the purposes, and an augmentation or dynamic feature lifting is required to incorporate sensor measurement functions. In this regard, we define the augmented state $\boldsymbol{S}$ as a feature vector that encompasses such parts:
\begin{equation}
    \boldsymbol{S}=\boldsymbol{g}(\boldsymbol{s})
\end{equation}

Numerous options exist for the mapping function $\bm{g}$, which can enhance sensor measurements and improve model accuracy. If the sensors are adequate to determine the system state, the identity map can be utilized as $\bm{g}$, which means $\bm{S}=\bm{s}$. Alternatively, $\bm{g}$ can leverage Proper Orthogonal Decomposition (POD) mode coefficients when the measurements provide high-dimensional snapshots. \citet{takensDetectingStrangeAttractors1981} and \citet{bruntonChaosIntermittentlyForced2017} use delay embedding technology to augment the measurements, resulting in a sufficiently high-dimensional feature vector that fully characterizes the system dynamics. Selecting an effective transformation function $\boldsymbol{g}$ is a critical unresolved issue relevant to both representation theory and the Koopman operator viewpoint on dynamical systems. Both \citet{mezicSpectralPropertiesDynamical2005} and \citet{bruntonKoopmanInvariantSubspaces2016} are actively investigating this problem. In this study, we choose g to augment the sensor measurement with its time derivative while appropriately scaling the augmented measurement. Furthermore, \citet{loiseauSparseReducedorderModelling2018} propose a comprehensive sparse reduced-order modeling for flow full-state estimation, which includes time-resolved sensor data and optional non-time-resolved particle image velocimetry (PIV) snapshots.
Inspired by the facts as mentioned above, we present a novel approach, named dynamic feature-based DRL, to overcome the limitations of measurements in the real world and highlight the potential of deep reinforcement learning techniques for sparse surface pressure sensing. An effective augmentation function $\boldsymbol{g}$ at time $t$ is used to lift the sensor signals so that a high-dimensional dynamic feature space is formed, which can be expressed as
\begin{equation}
    \boldsymbol{S}_t = \left( {\begin{array}{*{20}{c}}
  {\alpha s_{t - M}^1}& \cdots &{\alpha s_{t - M}^i}&{\beta a_{t - M}^1}& \cdots &{\beta a_{t - M}^j} \\ 
   \vdots & \ddots & \vdots & \vdots & \ddots & \vdots  \\ 
  {\alpha s_t^1}& \cdots &{\alpha s_t^i}&{\beta a_t^1}& \cdots &{\beta a_t^j} 
\end{array}} \right) \in\mathbb{R}^{(M+1)\times (i+j)},
\end{equation}
where $a$ is the agent action at time $t$, $M$ is the number of backtracking time steps $t$, which is set to 30 in this study, corresponding to twice time steps of the baseline vortex shedding period, $i$ and $j$ are the identifiers of sensor and actuator, respectively, $\alpha$ and $\beta$ are corresponding scaling factor. The final algorithm is listed in Algorithm \ref{alg:DF-DRL}. In the context of DF-DRL with SAC, the approach involves a cyclic iteration of gathering experience from the environment based on the current policy and updating the function approximators by utilizing stochastic gradients from batches sampled from a replay pool. The original state vector within a state-action tuple will first be lifted according to the augmentation function $g$, then dumped to the replay buffer $\mathcal{D}$.

\SetKwComment{Comment}{/* }{ */}
\SetKwInOut{init}{Initialization}

\begin{algorithm}[htb]
\caption{The DF-DRL algorithm with Soft Actor-Critic}\label{alg:DF-DRL}
\KwIn{sensor state vector $s \in\mathbb{R}^{1\times i}$, initail critic network parameters $\theta_i$, initail actor network parameters $\phi$, state transition probability $p$ (deterministic CFD simulation in this study)}
\init{target network weights $\overline{\theta_i} \gets \theta_i$, replay buffer $\mathcal{D} \gets \emptyset$}
\For{each iteration}{
    \For{each environment step}{
        $a_t \sim \pi_\phi\left(a_t \mid s_t\right)$\;
        $s_{t+1} \sim p\left(s_{t+1} \mid s_t, a_t\right)$\;
        $\boldsymbol{S}_t \gets \boldsymbol{g}(\{s_1,s_2,\dots,s_{t+1}\} \cup \{a_1,a_2,\dots,a_{t}\})$\;
        $\mathcal{D} \leftarrow \mathcal{D} \cup \left\{(\boldsymbol{S}_t, a_t, r(s_t, a_t))\right\}$\;
    }
    \For{each gradient step}{
        Update $\theta_i$, $\phi$, $\overline{\theta_i}$, and temperature $T$ according to \citep{haarnoja2018soft}\;
    }
}
\KwOut{actuator control vector $a \in\mathbb{R}^{1\times j}$}
\end{algorithm}

\begin{figure}
    \centering
    \begin{overpic}[width=0.9\textwidth]{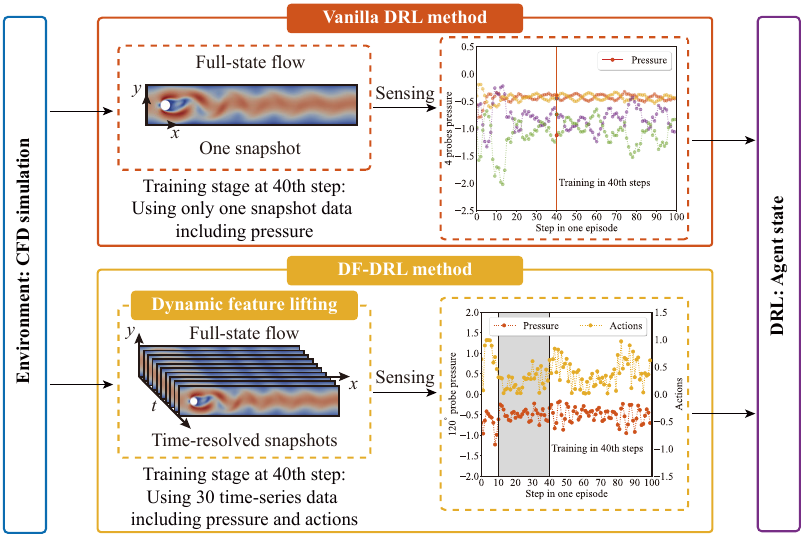}
        \put(36.5,64){\textcolor{white}{\fig{a}}}
        \put(38,32.5){\textcolor{white}{\fig{b}}}
    \end{overpic}
\caption{Flowchart of the two approaches for the state collected from the environment. \fig{a} Vanilla method (Sensor-feedback): the agent only collects the flow field state at a single time step. For example, the signal obtained from four pressure sensors located in the flow field at the 40th time step returns a state vector $s\in\mathbb{R}^{4\times 1}$; \fig{b} DF-DRL method: the agent collects data from the most recent thirty time steps $t$, including historical sensor pressure $p\in\mathbb{R}^{30\times 1}$ and action data $a\in\mathbb{R}^{30\times 1}$ provided by the agent. This process indicates dynamic feature lifting and the dimension of the state vector $\boldsymbol{S}\in\mathbb{R}^{30\times 2}$. Moreover, scaling the state vector will amplify signal fluctuations, which is helpful in capturing the flow characteristics.}
\label{fig:df-drl-sensing}
\end{figure}

In past studies, a common practice has been to use a single snapshot of the flow field as state data, such as four sensors of pressure data in one time step, to provide input to the policy network. This is illustrated by the upper panel of Fig. \ref{fig:df-drl-sensing}. By contrast, the DF-DRL method uses pressure data assembled from sensor measurements extracted from the 30 previous action time steps $t$, resulting in an augmented agent state. The details of dynamic feature lifting within the DF-DRL method are illustrated in lower Fig. \ref{fig:df-drl-sensing}. More specific DF-DRL hyperparameters have been listed in Table \ref{tab:Hyperparameters_DF-DRL} of Appendix \ref{sec:hyperparameters}. It is expected (and confirmed in \S \ref{sec:DF-DRL_Results}) that the policy will be improved using such dynamic feature lifting input data. However, a possible challenge is that this may increase the dimensionality of the state input to the ANN quite a bit since it is a two-dimensional array with one dimension corresponding to the sensor number and another corresponding to the time series index. In particular, sensors on the surface observe lower magnitude variations in flow velocity and pressure than sensors in the wake. They cannot observe changes in the trend of the wake and the shedding of cylinder vortices. Therefore, using the DF-DRL method is most appropriate for the surface sensors AFC training process, which involves fewer sensors. Besides, it is also vital to use an input standardization method individually on each sensing time series. In particular, it is necessary to normalize the surface pressure sensor observations so that these fluctuations are well perceived even though these have a very different dynamic range compared with sensors in the wake region.

\section{Numerical plant: flow around a circular cylinder} \label{sec:Plant}
In this section, we choose benchmarks proposed by \citet{schafer1996benchmark}, laminar flow around a cylinder, and place jets symmetrically arranged on both lateral sides as actuators for active flow control. The objective is to reduce the cylinder's drag force and lift fluctuation. Firstly, we formalize the research problem in \S \ref{sec:formalization}. Then, we provide a detailed description of the flow configuration and numerical solution methods in \S \ref{sec:CFD}, followed by a validation of the accuracy of the numerical algorithms. Finally, we define three types of sensor layout schemes in \S \ref{sec:Layout}.

\subsection{Problem formalization} \label{sec:formalization}
The active flow control task formulated in this study aims to find a real-time control policy $\displaystyle \pi $ of two jet actuators located on a circular cylinder with sensor feedback, which can effectively reduce the fluid force on it. Generally, the surface pressure information $\displaystyle s_{t}$ can be regarded as the input state of the control policy $\displaystyle \pi $, and the jet intensity can be viewed as the DRL action $\displaystyle a_{t}$ at the time $\displaystyle t$. The action is decided by the DRL controller based on the state observation. Therefore, the control processing can be modeled as a deterministic or stochastic relationship:
\begin{equation}
a_{t} \ \sim \ \pi ( s_{t} |\theta ) .
\end{equation}
Hence, given a deep reinforcement learning agent with the control policy $\displaystyle \pi $, the objective is to minimize the lift and drag coefficients of the cylinder by optimizing the set of weights $\displaystyle \theta $ of the DRL agent policy network:
\begin{gather}
\pi ^{*} =\pi \left( \theta ^{*}\right) \label{eq:optimal_policy}, \\
\theta ^{*} =\arg\max_{\pi ( \theta )}\mathbb{E}_{( s_{t} ,a_{t}) \sim \rho _{\pi ( \theta )}}\mathcal{T}( s_{t} ,a_{t}),
\label{eq:optimal_args}
\end{gather}
where the superscript * represents the optimal value, $\mathbb{E}$ is the expected value operator, and $\displaystyle \mathcal{T}$ denotes a target function, which represents the current policy $\pi$.

\subsection{Flow configuration and numerical method} \label{sec:CFD}
In this work, we use the open-source CFD package OpenFOAM to perform simulations. Under the assumption of incompressible viscous flow, the governing Navier-Stokes equations can be expressed in a non-dimensional manner as:
\begin{gather}
\frac{\partial \boldsymbol{u}}{\partial t}+\boldsymbol{u} \cdot(\nabla \boldsymbol{u})=-\nabla p+\operatorname{Re}^{-1} \Delta \boldsymbol{u}, \\
\nabla \cdot \boldsymbol{u}=0, \\
Re = \frac{\bar{U}D}{\nu},
\label{con:govern}
\end{gather}
where $\boldsymbol{u}$ is the non-dimensional velocity, $t$ is the non-dimensionless time, $p$ is the non-dimensional pressure, $\nu$ is the kinematic viscosity of the fluid, and $\bar{U}$ is the mean velocity at the inlet. The corresponding Reynolds number $Re$ is 100 in the training stage.

\begin{figure}
\centering
    \begin{overpic}[width=0.8\textwidth]{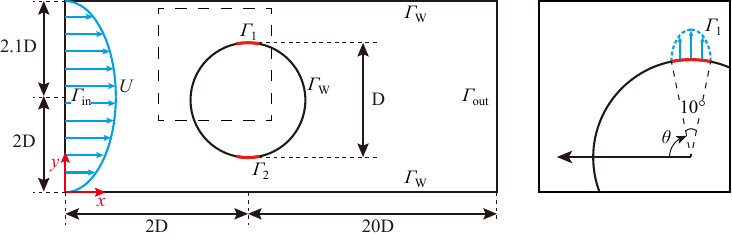}
        \put(0.5,29.5){\fig{a}}
        \put(74.5,29.5){\fig{b}}
    \end{overpic}
\caption{Description of \fig{a} the numerical setup, which is adapted from \citet{schafer1996benchmark}. The origin of the coordinates is located at the lower left corner of the entire computational domain. $\Gamma_{in}$ stands the inflow velocity with a parabolic flow profile, while $\Gamma_{out}$ is put for the outflow. Non-slip wall boundary constraint $\Gamma_W$ is applied on the bottom and top of the channel and on the cylinder surfaces. Two jet holes ($\Gamma_1$ and $\Gamma_2$) are present at both sides of the cylinder; \fig{b} An enlarged view of the dashed box in subfigure \fig{a}. The $0^\circ$ azimuth angle corresponds to the foremost point on the cylinder's windward surface and increases clockwise. The jet actuator opening angle is $10^\circ$, consistent with \citet{Rabault2019Artificial}.} 
\label{fig:flow_conf}
\end{figure}

This study focuses on two-dimensional simulations of flow around a circular cylinder with a diameter $D$, which is the characteristic length scale. The computational domain has dimensions of $L = 22D$ and $B = 4.1D$ in the streamwise and cross-stream directions, respectively, as shown in Fig. \ref{fig:flow_conf}. Following the widely recognized benchmark conducted by \citet{schafer1996benchmark}, the cylinder is slightly off-center to induce faster development of the vortex shedding alley during the initial simulation convergence stage. The outlet boundary is placed $19.5D$ downstream of the cylinder to allow the wake to develop fully. 

The inlet boundary denoted as $\Gamma_{in}$ is subject to the parabolic velocity inlet boundary condition. The no-slip constraint, $\Gamma_{W}$, is applied to both the top and bottom of the channel and the surfaces of the cylinder. Additionally, the right boundary of the channel, $\Gamma_{out}$, is designated as a pressure outlet, wherein zero velocity gradient and constant pressure are maintained. The inlet boundary is assigned as a parabolic velocity form and expressed as the following in the streamwise direction:
\begin{equation}
        U(0, y)=4 U_{m} y(H-y) / H^{2},
\label{con:inletsudu}
\end{equation}
where $U_{m}$ is the maximum inflow velocity at the middle of the channel. Employing a parabolic inflow profile, $U_{m}$ is 1.5 times the mean velocity $\bar{U}$, as defined by:
\begin{equation}
\bar{U}=\frac{1}{H} \int_{0}^{H} U(y) \mathrm{d} y=\frac{2}{3} U_m .
\label{con:um}
\end{equation}

To accomplish the active flow control, the flow control technique using two jet actuators ($\Gamma_1$ and $\Gamma_2$) located on opposite sides of the cylinder is employed. A parabolic velocity profile with a jet width of $\omega = 10^{\circ}$ is imposed at both jets, as depicted in Fig. \ref{fig:flow_conf}. Due to the velocity of the jet flow being orthogonal to the inflow direction, drag reduction is strictly achieved by effective actuation rather than by momentum injection. Moreover, the jet flow on both sides is constrained as synthetic jet flow, i.e., $V_{\Gamma_1} = V_{\Gamma_2}$, so the jets collectively do not add or remove mass to the flow. The normalized jet flow rate $Q_i^*$ of a jet is defined as
\begin{equation}
Q_i^* = \frac{Q_{jet,i}}{Q_{ref}} = \frac{U_{jet,i}\cdot D_{jet}}{\bar{U}  D} \label{flowrate}
\end{equation}
where $U_{jet,i}$ is the $i_{th}$ jet velocity, $D_{jet}$ is the width of the jet, and $Q_{ref}$ is the reference mass flow rate intercepting the cylinder. $Q_i^*$ is not greater than 0.2 during this study.

The current study adopts unstructured meshes for computational fluid dynamics (CFD) simulations. Emphasis has been laid on refining the mesh around the surface boundary and the wake flow regions, as these are crucial for ensuring the appropriate resolution of these significant flow domains and the physics happening there. The numerical solution is obtained at each time step, and the drag ($F_D$) and lift ($F_L$) forces are computed by integrating over the cylinder surface, following:
\begin{equation}
\boldsymbol{F}=\int(\boldsymbol{\sigma } \cdot \boldsymbol{n}) \cdot \boldsymbol{e}_{j} \mathrm{~d}  ,
\label{eq:intergral_force}
\end{equation}
where $\sigma$ is the Cauchy stress tensor, and the unit vector $n$ is defined as normal to the cylinder surface. At the same time, $e_j$ is denoted as a unit vector in the direction of the inflow velocity for drag force calculations and as a vector perpendicular to the inflow velocity for lift force calculations. Specifically, the drag $C_D$ and lift $C_L$ coefficients can be expressed as follows:
\begin{gather}
C_{D}=\frac{F_{D}}{\frac{1}{2} \rho \bar{U}^{2} D} \label{eq:cd}, \\
C_{L}=\frac{F_{L}}{\frac{1}{2} \rho \bar{U}^{2} D} \label{eq:cl},
\end{gather}
where $F_D$ and $F_L$ are denoted as integral drag and lift force, respectively.

To further validate the accuracy of the CFD simulations, a series of mesh convergence studies are performed at a Reynolds number $Re = 100$. In particular, meshes of three different resolutions are employed. The corresponding results for the maximum values of the drag coefficient ($C_D$) and lift coefficient ($C_L$), denoted as $C_{D}^{\max}$ and $C_{L}^{\max}$, respectively, are reported in Table \ref{tab:cylinder_Re}. The numerical analysis reveals that the discrepancies among various mesh resolutions are insignificant. Considering the trade-off between computational cost and numerical accuracy, the meshing scheme of Grid II is preferred for the DRL training stage. More specific flow configurations have been listed in Table \ref{tab:Hyperparameters_flow} of Appendix \ref{sec:hyperparameters}.

\begin{table}
    \centering
    \caption{Numerical results of mesh convergence study for the 2D flow around a circular cylinder at $Re = 100$.} \label{tab:cylinder_Re}
        \begin{tabularx}{\textwidth}{P{10em}CCCCC}
            \toprule
            \textbf{Case} & \textbf{Grid number} & \textbf{$C_{D}^{\max }$} & \textbf{$C_{L}^{\max }$} & \textbf{$\overline{C_{D}}$} & \textbf{$St$} \\
            \midrule
            Grid I   & 13572 & 3.27 & 1.01  & 3.21  & 0.301 \\ 
            Grid II  & 16200 & 3.24 & 1.024 & 3.205 & 0.299 \\
            Grid III & 39382 & 3.24 & 1.04  & 3.21  & 0.302 \\
            \citet{schafer1996benchmark} &  & 3.22-3.24 & 0.99-1.01 &  & 0.295-0.305 \\
            \citet{Rabault2019Artificial} & 9262 & 3.245 & 1.020 &  & 0.302 \\
            \bottomrule
        \end{tabularx}
\end{table}

\subsection{Layout of surface pressure sensors} \label{sec:Layout}
\begin{figure}
\centering
\begin{overpic}[width=0.97\textwidth]{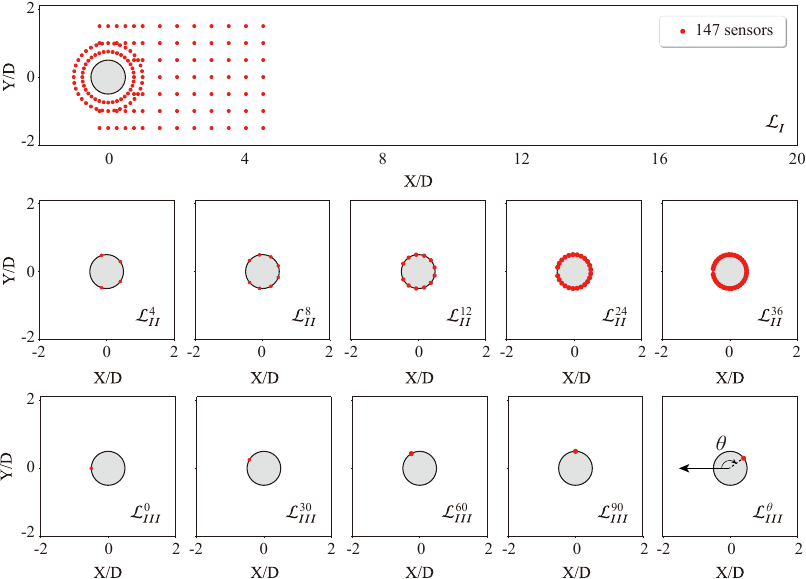}
    \put(-3,70.5){\fig{a}}
    \put(-3,46){\fig{b}}
    \put(-3,22){\fig{c}}
\end{overpic}
\caption{Number and configuration of the sensors used to generate the DRL controller state observation: \fig{a} Using 147 sensors provides sufficient flow information for DRL training, stated as \layout{1}{}; \fig{b} Layout using 4, 8, 12, 24, and 36 sensors in the surface of the cylinder respectively, denoted by \layout{2}{4}, \layout{2}{8}, \layout{2}{12}, \layout{2}{24}, and \layout{2}{36}; \fig{c} Layout using only one sensor located on the surface of the cylinder with an azimuth angle of $0^{\circ}$, $30^{\circ}$, $60^{\circ}$, $90^{\circ}$, and $\theta^{\circ}$, signified by \layout{3}{0}, \layout{3}{30}, \layout{3}{60}, \layout{3}{90}, and \layout{3}{\theta}.} \label{fig:sensor_location}
\end{figure}

The present study proposes a series of pressure sensor layout schemes to study the influence of sensor location. First, a baseline configuration proposed by \citet{Rabault2019Artificial} with 147 pressure sensors is set up both around the cylinder and in the wake region, as shown in Fig. \ref{fig:sensor_location}\fig{a}. Then, A varying number of pressure sensors, \eg 4, 8, and 24 sensors, are symmetrically arranged on the surface of the cylinder (along the direction of inflow). To avoid inadequate information with more minor pressure fluctuations at the front of the cylinder, the sensors are uniformly distributed except for the point at the front, as shown in Fig. \ref{fig:sensor_location}\fig{b}. Finally, a comprehensive study is carried out using a single sensor location. The placement of the single sensor is started by putting it at the front of the cylinder as $\theta = 0^\circ$ relative to the incoming flow. We change its position by gradually increasing its angular position on the cylinder in increments of $15^\circ$ until it reaches the rear edge of the cylinder ($\theta = 180^\circ$), as shown in Fig. \ref{fig:flow_conf}\fig{c} and Fig. \ref{fig:sensor_location}\fig{c}. As a consequence, a total of 17 single pressure sensor positions are investigated. One could expect that pressure sensors on the surface of the cylinder can provide valuable information about the flow to the DRL controller. However, using surface sensors like those shown in Fig. \ref{fig:sensor_location}\fig{b} and \fig{c} presents a challenge due to the limited quantity of data provided and the placement being solely on the surface of the cylinder. This results in insufficient information regarding the cylinder wake and vortex shedding pattern during the DRL training stage.

To facilitate the description in the next sections, the notation $\mathcal{L}$ is used to describe the different sensor layout schemes. The subscripts represent different sensor layout types. For example, \layout{1}{} represents the baseline configuration with 147 sensors placed inside the flow field, \layout{2}{} represents the sensor configuration placed on the cylinder surface, and \layout{3}{} represents a single sensor configuration placed on the cylinder surface. For type \layout{2}{N}, the superscript $N$ indicates the number of sensors, and the polar coordinates of sensor $i$ can be expressed as
\begin{equation}
    r_i = \frac{1}{2}D, \quad \theta_i = \frac{2\pi i}{N+1}, \quad i=1,2,\dots,N,
\end{equation}
while for type \layout{3}{\theta}, the superscript $\theta$ indicates the degree of the sensor placement where the coordinate axis of the polar coordinate system with origin opposite to the inflow direction, \ie the leading edge of the cylinder is denoted as the $0^\circ$ point, which is illustrated in Fig. \ref{fig:sensor_location}.

\section{Results and discussion} \label{sec:Results}
In this section, we first evaluate the performance and reliability of the proposed DF-DRL approach in \S \ref{sec:DF-DRL_Results}, comparing it with a vanilla DRL algorithm. Then, the impact of different numbers of surface sensor configurations and layouts of single surface sensors on the performance of flow control are investigated in \S \ref{sec:sensor_performance}. Furthermore, we verify the robustness of the DF-DRL controllers under two different Reynolds numbers, $Re=500$ and $Re=1000$ in \S \ref{sec:robustness}.

\subsection{DRL-based AFC with sparse surface pressure sensing} \label{sec:DF-DRL_Results}
\begin{figure} \centering
    \begin{overpic}[width=0.48\textwidth]{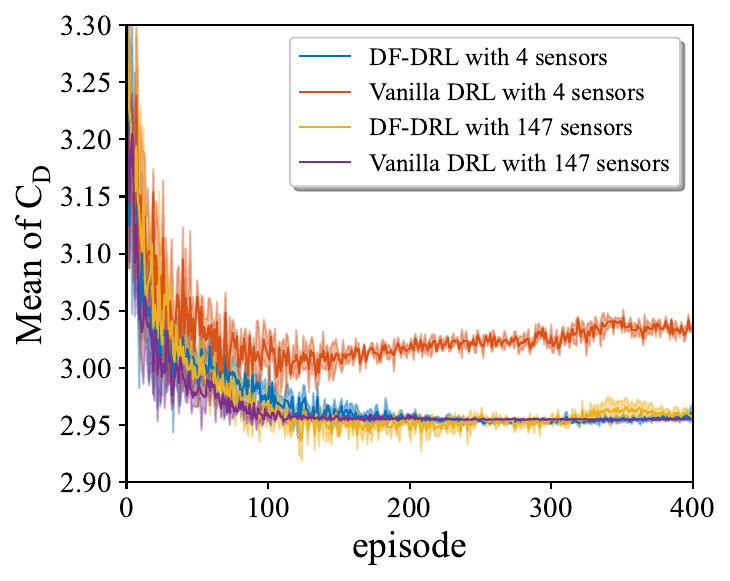}
        \put(0,75){\fig{a}}
    \end{overpic}
    \hspace*{\fill}
    \begin{overpic}[width=0.48\textwidth]{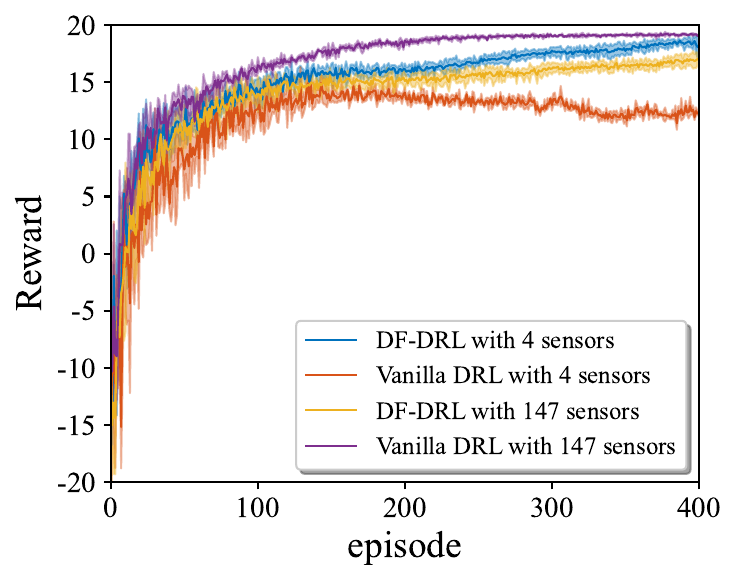}
        \put(0,75){\fig{b}}
    \end{overpic}
    \\
    \begin{overpic}[width=0.48\textwidth]{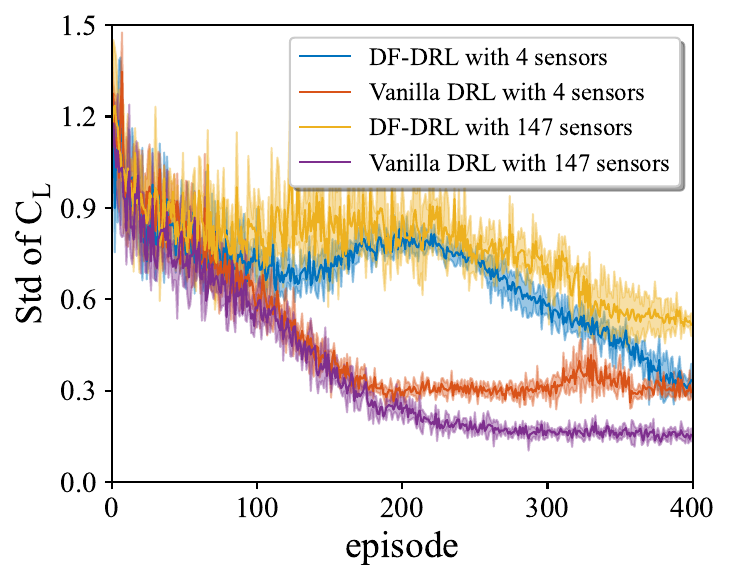}
        \put(0,75){\fig{c}}
    \end{overpic}
    \caption{Comparison of \fig{a} the mean $C_D$, \fig{b} reward, and \fig{c} the std of $C_L$ when using different DRL methods, \ie using the DF-DRL method or not, and a different number of the sensor (4 sensors in the surface of cylinder and 147 sensors around the cylinder). The learning case condition, which contains 147 sensors without time series, reaches the maximum reward. These cases are trained three times repeatedly, and we present the average between the three runs (thick line) and the std between these runs (shadowed area).} \label{fig:comparison_DF-DRL}
\end{figure}

To evaluate the effectiveness and reliability of the DF-DRL approach, sensor locations \layout{1}{} and \layout{2}{4} are selected for illustration, as depicted in Fig. \ref{fig:sensor_location}\fig{a} and \fig{b}, respectively.  Figure \ref{fig:comparison_DF-DRL} shows the learning curves for active flow control using both vanilla DRL and DF-DRL techniques under two different sensor quantity configurations (4 and 147 sensors). The three subplots \fig{a}, \fig{b}, and \fig{c} correspond to the mean $C_D$, reward, and the standard deviation (std) of $C_L$, respectively. Although it seems that the $C_L$ learning curve, as shown in Fig. \ref{fig:comparison_DF-DRL}\fig{c}, has not reached a converged state, especially for the two cases that utilize the dynamic feature lifting technique. However, we still believe that the DRL learning has up to steady according to the clear convergence of the $C_D$ learning curve, as depicted in Fig. \ref{fig:comparison_DF-DRL}\fig{a}. This is because, as explained in \S \ref{sec:DRL}, $C_L$ is only used as a penalty term to prevent the policy network from finding control strategies that are not practical or realistic.

The results indicate that there are significant differences between these four cases. Using a vanilla DRL algorithm, the maximum drag reduction of 8$\%$ was achieved with the use of 147 sensors (scheme \layout{1}{}), and the maximum reward value is 19.13. Moreover, the std of $C_L$ is reduced to 0.15 when learning has converged. For the case of 4 sensors with vanilla DRL method (scheme \layout{2}{4}), the maximum drag reduction is only 6.4\%, and the std of $C_L$ decreases to only 0.29, as shown in Fig. \ref{fig:comparison_DF-DRL}. These two cases show that with few surface sensors, active flow control performance worsens. This is due to the inability of DRL to estimate the flow field and limited observable data correctly.

As for the results using the DF-DRL method, it is observed that both \layout{1}{} and \layout{2}{4} achieve similar drag reduction amplitudes of 8\%, corresponding to the maximum drag reduction also observed using the 147 pressure sensors. However, the latter approach achieves a higher reward value, as indicated by the decrease in the std of the $C_L$ due to the lift penalty term within the reward function. This performance improvement is happening even though both the \layout{1}{} and \layout{2}{4} approaches undergo a temporary rebound in their learning. Therefore, when utilizing the DF-DRL method, fewer sensors, as used in scheme \layout{2}{4}, can achieve the same drag reduction as scheme \layout{1}{}, which achieves $25\%$ better than the vanilla model based on direct sensor-feedback. This can be obtained while also improving the reduction of lift fluctuations. These observations suggest that the DF-DRL method can maintain drag-reduction performance while reducing the number of sensors required. The difference in reward observed between the two cases can be attributed to the reduction in the std of $C_L$ achieved in scheme \layout{2}{4}.

For scheme \layout{1}{}, with many sensors distributed around the cylinder and wake region, the DRL controller can obtain exhaustive flow field information. Thus, the inclusion of historical data provided by the DF-DRL method has a minor impact. However, scheme \layout{2}{4}, with limited sensor numbers and sparse information on the cylinder surface, has more information available to perform effective flow control with the DF-DRL method. These results suggest, unsurprisingly, that using more sensors leads to better $C_D$ reduction effects and more stable reward convergence with naive DRL agents. Moreover, when the flow field information is limited in quantity and placement of the sensors (due to physical restrictions), the DF-DRL method demonstrates better convergence and yields a superior control policy.

\subsection{Control performance and learning convergence with DF-DRL method} \label{sec:sensor_performance}
To further investigate the impact of sensor quantity and placement azimuth on the control effectiveness and convergence performance of the DF-DRL-based controller, we conducted case studies with different quantities of the sensor, \ie 1, 4, 8, 12, 24, and 36, as well as various placement azimuth layouts of $0^\circ$ to $180^\circ$ with a $15^\circ$ spacing.

\subsubsection{Sensor quantity}
Five typical layout schemes \layout{2}{} of surface pressure sensors are investigated in this section. The arrangement of these sensors is depicted in Fig. \ref{fig:sensor_location}\fig{b}, where the sensors are evenly distributed, and all the leading edge sensor is removed, as described in \ref{sec:Layout}. Since the state includes the jet actions component, the pressure sensors distributed around the jet will not significantly impact the results.

\begin{figure} \centering
    \begin{overpic}[width=0.48\textwidth]{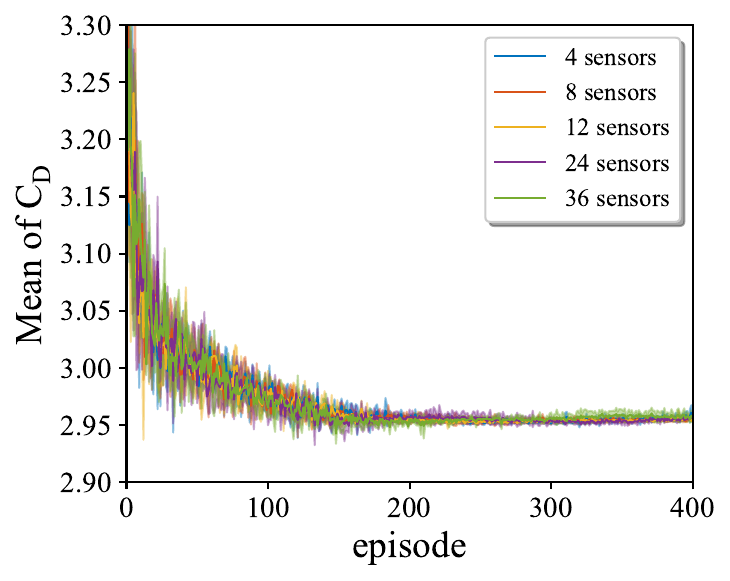}
        \put(0,75){\fig{a}}
    \end{overpic}
    \hspace*{\fill}
    \begin{overpic}[width=0.48\textwidth]{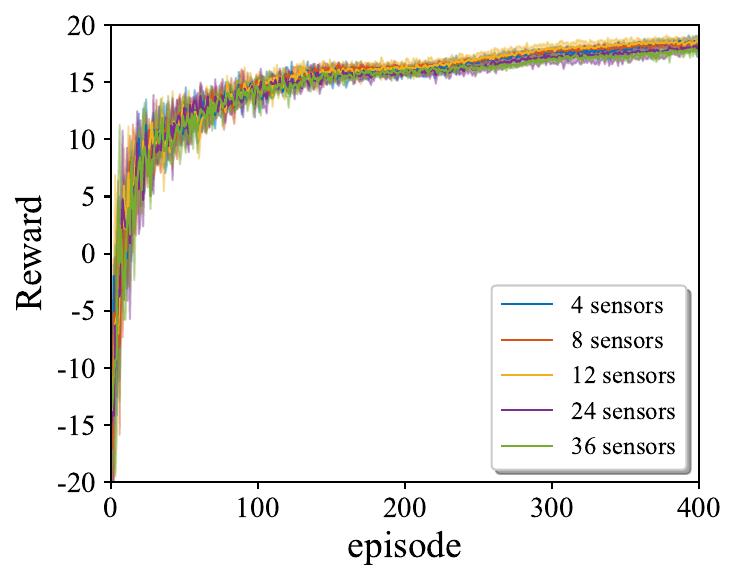}
        \put(0,75){\fig{b}}
    \end{overpic}
    \\
    \begin{overpic}[width=0.48\textwidth]{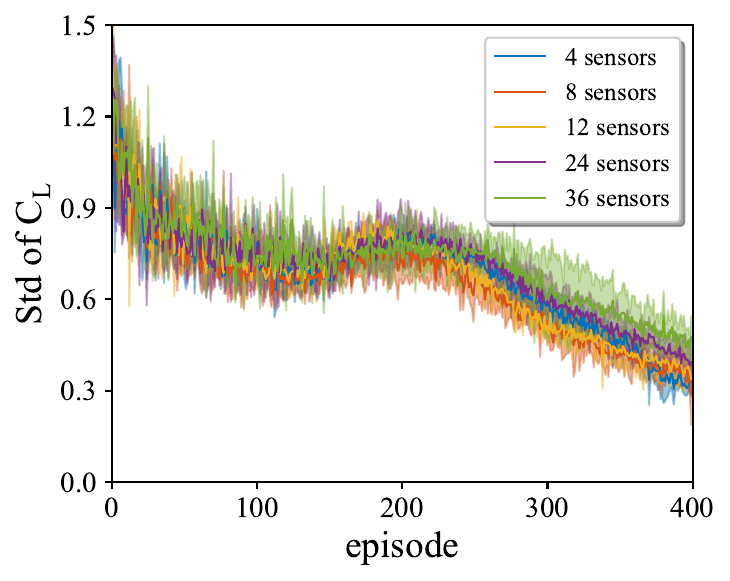}
        \put(0,75){\fig{c}}
    \end{overpic}
    \caption{Evolution of \fig{a} mean $C_D$ reduction, \fig{b} reward, and \fig{c} std lift coefficient in the different number of sensors with DF-DRL method. Sensors are uniformly located on the surface of the cylinder as indicated in Fig. \ref{fig:sensor_location}\fig{c}.} \label{fig:sensor_number}
\end{figure}

The impact of adding more surface pressure sensors on training performance is illustrated in Fig. \ref{fig:sensor_number}. Results show that increasing the number of sensors does not lead to a significant improvement in drag and $C_L$ reduction, which remains around 8\% across all schemes. Additionally, all cases converge at approximately 200 episodes. Figure \ref{fig:sensor_number}\fig{b} displays learning curves that follow the same trend as the $C_D$, indicating that the final DF-DRL performance is very similar to the benchmark case for different numbers of pressure sensors on the surface of the cylinder.

As depicted in Fig. \ref{fig:sensor_number}\fig{c}, the standard deviation of the lift exhibits a comparable declining pattern to that observed in the previous section. Following an initial decrease and temporary increase at around episode 200, all five groups undergo a consistent decrease until the completion of DF-DRL training.

This result can be explained by the fact that the pressure on the cylinder surface, as a surrogate for lift, is a better dynamic feature than wake measurements, where varying vortex shedding destroys the phase relationship. The pressure on the cylinder surface provides a more accurate representation of the dynamic behavior of the flow, leading to a better understanding of the underlying physics. This conclusion agrees with previous work by \citet{loiseauSparseReducedorderModelling2018}. Using pressure as a dynamic feature highlights the potential for feature-based approaches in reduced-order modeling of fluid flows.

Even though the final drag reduction effect is very similar, an interesting subtle difference can be observed from Fig. \ref{fig:sensor_number}, where the DRL algorithm with dynamic feature lifting does not show a monotonically increasing relationship between the number of sensors and control performance.

The counterintuitive scenario mainly arises from the potential redundant information (features) in the state \citep{roghairVisionBasedDeep2022a}. This can be explained by the phenomenon of the curse of dimensionality \citep{altmanCurseDimensionality2018} commonly observed in deep learning. The curse of dimensionality refers to a series of problems and challenges encountered in high-dimensional spaces. It describes the difficulties and obstacles faced when dealing with data with a large number of dimensions.

The main issues associated with the curse of dimensionality in high-dimensional spaces include (1) the distances between data points become very large, leading to data sparsity. This means that it becomes challenging to find other similar data points near a given data point, making data analysis and modeling difficult; (2) selecting and extracting meaningful features becomes complex. Due to the large number of dimensions, it is difficult to determine which features are most important for solving a specific problem; (3) statistical inference and model estimation in high-dimensional data become challenging. Due to data sparsity and computational complexity, obtaining reliable statistical results from limited data samples becomes difficult, leading to issues like overfitting or underfitting.

In this study, the data collected by each sensor can be regarded as a feature. For a case with $i$ sensors and two actuators, considering the commonly used dynamic feature lifting, which includes the past $M$ time steps of pressure and the magnitude of actuator jet velocity, the dimensionality increases from $\mathbb{R}^{1\times i}$ to $\mathbb{R}^{M\times (i+1)}$. When the dimensionality of the state increases, the complexity and variability of the data also increase. If there are too few sensors, important information and patterns may be missed, leading to suboptimal control performance. Insufficient sensors may result in incomplete data coverage or inadequate representation of the system dynamics, making it difficult to capture crucial features or relationships necessary for effective control.

On the other hand, if there are too many sensors, the curse of dimensionality comes into play. The high-dimensional space poses challenges in terms of data sparsity, computational complexity, and feature selection. The abundance of sensors may introduce noise, redundancy, or irrelevant information, which can hinder the control process. It becomes difficult to discern which sensors are truly informative and contribute meaningfully to the control objective.

Given that the disparities in $C_D$ and $C_L$ fluctuation reduction among the experiments are not substantial, to further demonstrate the potential of using surface pressure as a dynamic feature for a nonlinear system and its combination with DRL, we will reduce the number of sensors to one in the next section.

\subsubsection{Placement azimuth} \label{sec:results_angle}
\begin{figure} \centering
    \begin{overpic}[width=0.48\textwidth]{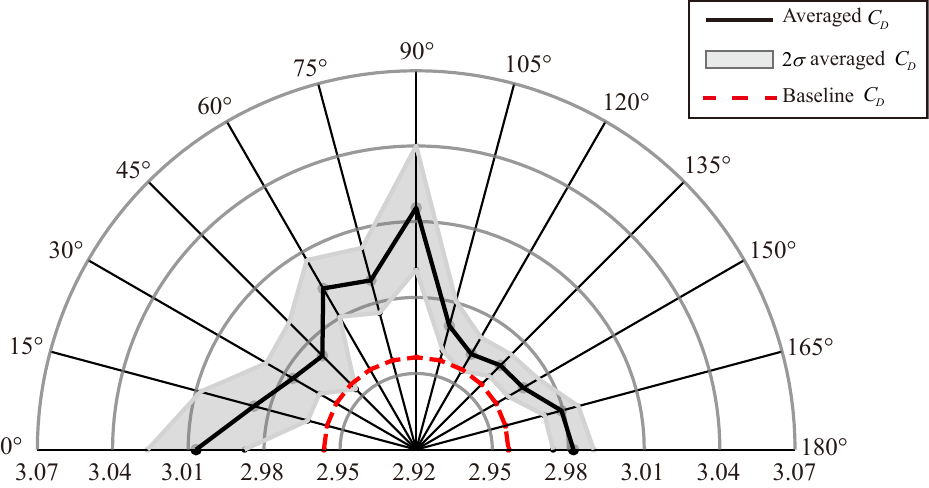}
        \put(0,40){\fig{a}}
    \end{overpic}
    \hspace*{\fill}
    \begin{overpic}[width=0.48\textwidth]{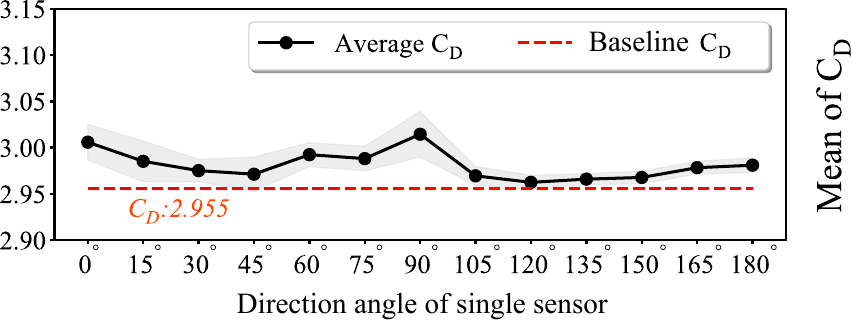}
        \put(-5,40){\fig{d}}
    \end{overpic}
    \\
    \begin{overpic}[width=0.48\textwidth]{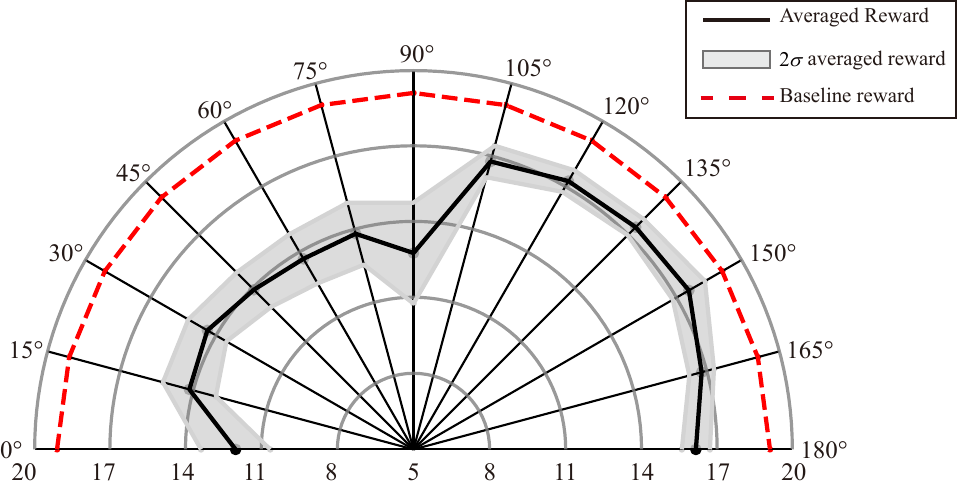}
        \put(0,40){\fig{b}}
    \end{overpic}
    \hspace*{\fill}
    \begin{overpic}[width=0.48\textwidth]{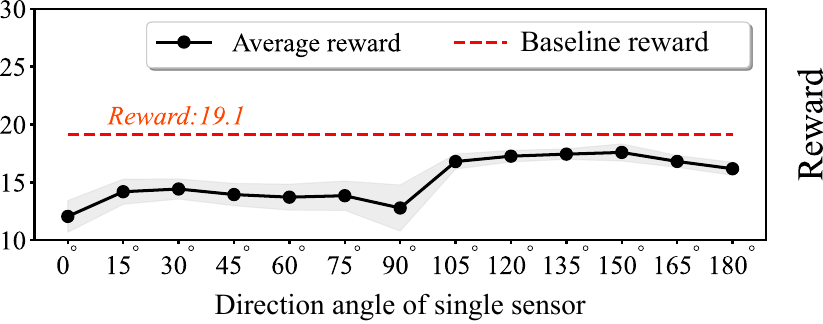}
        \put(-5,40){\fig{e}}
    \end{overpic}
    \\
    \begin{overpic}[width=0.48\textwidth]{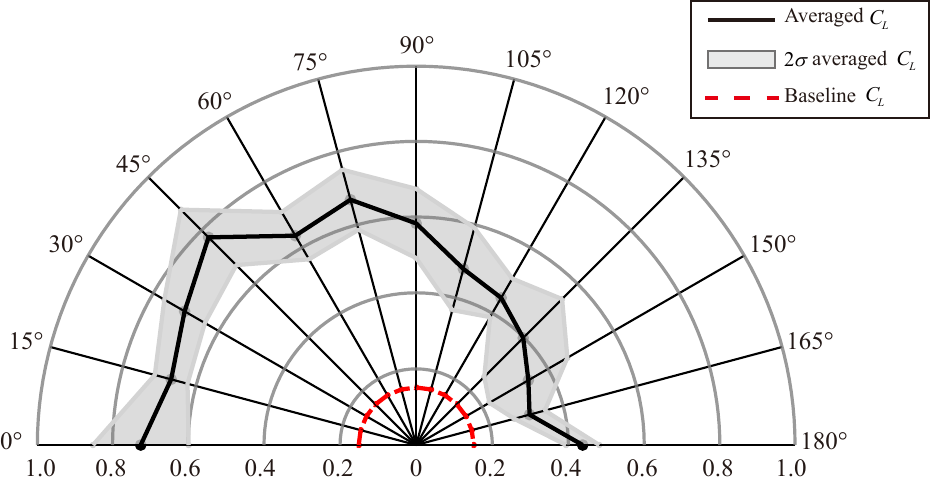}
        \put(0,40){\fig{c}}
    \end{overpic}
    \hspace*{\fill}
    \begin{overpic}[width=0.48\textwidth]{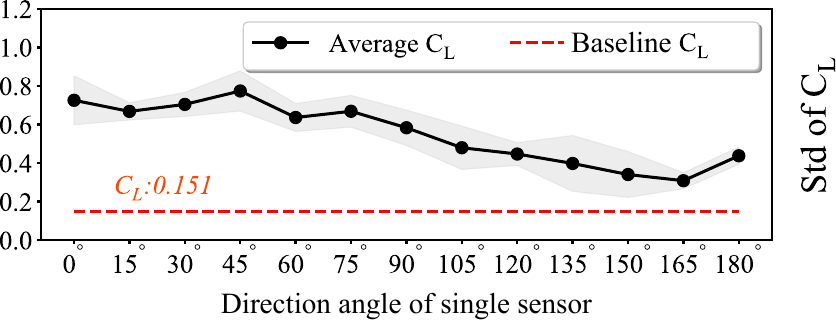}
        \put(-5,40){\fig{f}}
    \end{overpic}
    \caption{One pressure sensor is placed every $15^{\circ}$ on the cylinder surface for training, from $0^{\circ}$ to $180^{\circ}$, for a total of 13 sets of training. \fig{a}, \fig{b}, and \fig{c} are the mean $C_D$, reward, and the std of $C_L$ in the last 10 training episodes respectively, \fig{d}, \fig{e}, and \fig{f} is the changing trends of mean  $C_D$, reward and the std of $C_L$ respectively.} \label{fig:B_1}
\end{figure}

Based on the results above, it is apparent that an increase in the number of sensors utilized in DF-DRL-based AFC tasks does not necessarily result in better performance, including drag and lift reduction. To explore the maximum performance potential of DF-DRL and obtain the optimal sensor layout scheme, single sensor schemes \layout{3}{} are selected in the following study, as illustrated in Fig. \ref{fig:sensor_location}\fig{c}. According to the symmetry of the geometry of the setup and the boundary conditions, this study only considers deploying sensors on the upper semicircle region for training and analysis purposes. The cylinder surface features a single pressure sensor located every $15^{\circ}$, covering a range of $0^{\circ}$ to $180^{\circ}$, with a total of 13 configurations. Three training repetitions are performed for each case with the same hyperparameters to eliminate randomness. As shown in Fig. \ref{fig:B_1}\fig{a}, the mean $C_D$ indicates that AFC performance with only one pressure sensor is almost as optimal as baseline scheme \layout{1}{}. The results suggest that scheme \layout{3}{} on cylinder surfaces using the DF-DRL method can achieve the best active flow control performance.

It can also be observed from Fig. \ref{fig:B_1}\fig{b} that the trailing edge sensor of the cylinder has a higher reward value than the leading edge sensor, resulting in a lower mean $C_D$. Furthermore, a sudden reduction in the $C_D$ can be observed between \layout{3}{75} and \layout{3}{90}. This can be attributed to the influence of jet actuators situated on the top side of the cylinder, where changes in jet velocity can lead to abnormal pressure fluctuations on the surface at $90^{\circ}$ that confuse the DRL controller. Additionally, a significant jump in performance occurs between \layout{3}{90} and \layout{3}{150}, characterized by a marked decrease in the mean $C_D$. A decreasing trend can be observed in the std of $C_L$ from \layout{3}{0} to \layout{3}{180}, as depicted in Fig. \ref{fig:B_1}\fig{c}. This further emphasizes that a sensor located closer to the trailing edge of the cylinder can effectively generate information that can be used to mitigate the std of $C_L$.

\begin{figure}
\centering
\includegraphics[width=0.6\textwidth]{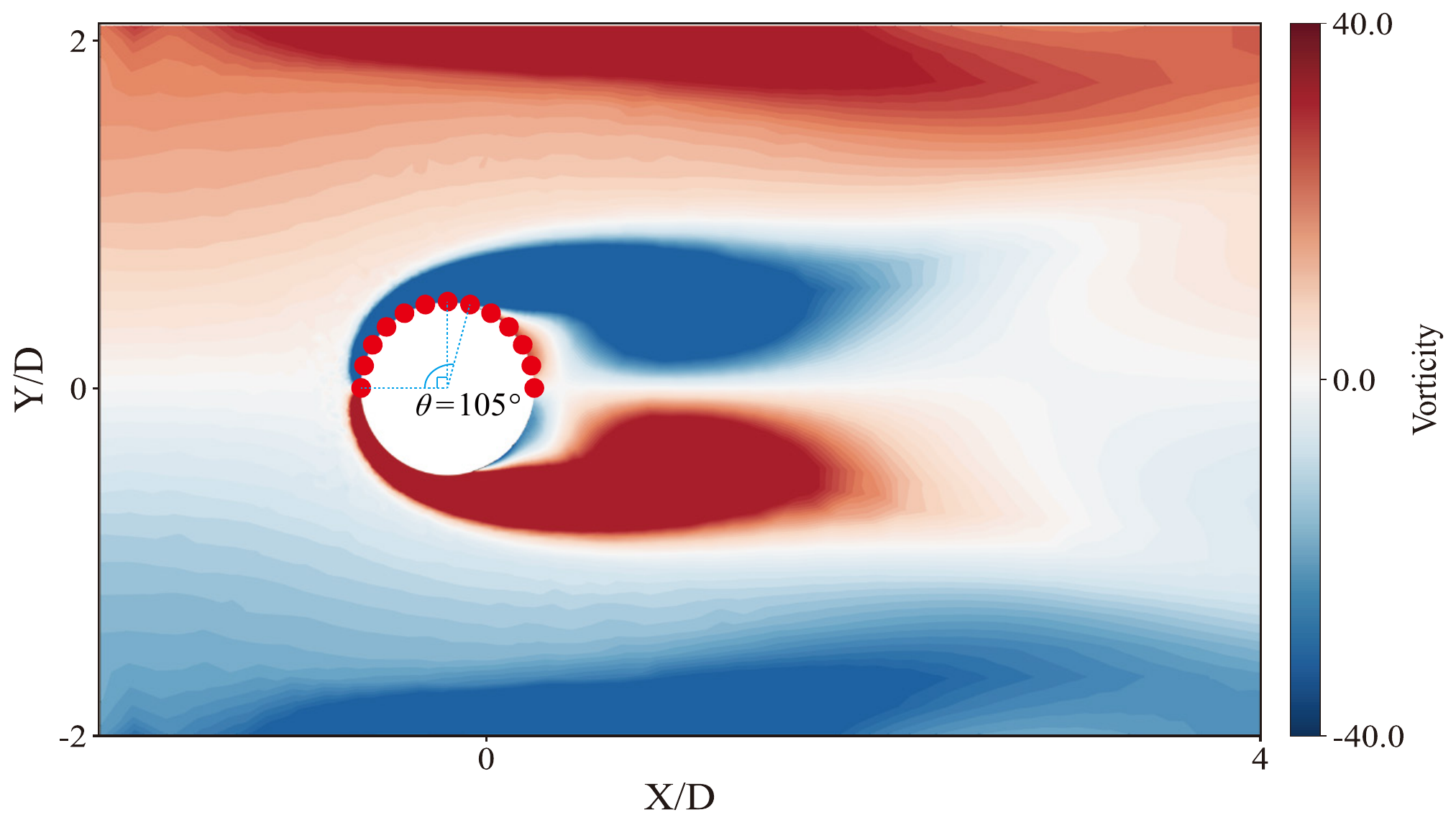}  
\caption{Contours of time-averaged vorticity of a plain case, the red points are surface pressure sensors. Vortex separation occurs between sensors $90^{\circ}$ and $105^{\circ}$.} \label{fig:100woliangtu}
\end{figure}

To explain this phenomenon more comprehensively, Fig. \ref{fig:100woliangtu} depicts the time-averaged vorticity field of the uncontrolled flow around the cylinder, with the red dots representing the 13 different single-sensor layout schemes. The sensors located at the trailing edge of the cylinder, namely \layout{3}{105} to \layout{3}{180}, are positioned at the vortex shedding location, indicating that these pressure sensors contain crucial information about vortex shedding compared to the windward side of the cylinder. This observation explains why the trailing edge pressure sensors outperform the leading edge sensors in the overall training outcomes, which include the reduction of $C_D$ and $C_L$.

To summarize, a general tendency of mean $C_D$, reward, and std of $C_L$ is presented in Fig. \ref{fig:B_1}\fig{d}. Notably, a single sensor situated between $0^{\circ}$ and $180^{\circ}$ demonstrates near optimal reduced $C_D$ and $C_L$, corresponding to an elevated reward value. These results inspire the training strategy of a single sensor system \layout{3}{} at higher Reynolds numbers.

\subsection{Robustness of DF-DRL based plant under more complex flow scenario} \label{sec:robustness}
\subsubsection{Case 1: 2D flow around a circular cylinder at higher Reynolds number}
Based on the promising performance of active flow control demonstrated in \S \ref{sec:results_angle}, Scheme \layout{3}{150} is chosen to investigate the robustness of a single surface sensor at higher Reynolds numbers ($Re$ = 500 and 1000). The policy network architecture is the same as before, consisting of two dense layers of 512 fully connected neurons, with the input layer receiving data from a single pressure sensor and the output layer representing the jet velocity. As the vortex shedding frequency of the cylinder increases with the rise in Reynolds numbers, the SAC agent interacts with the environment every 44 and 46-time step $t$ at $Re$ = 500 and 1000, respectively.

\begin{figure} \centering
    \begin{overpic}[width=0.45\textwidth]{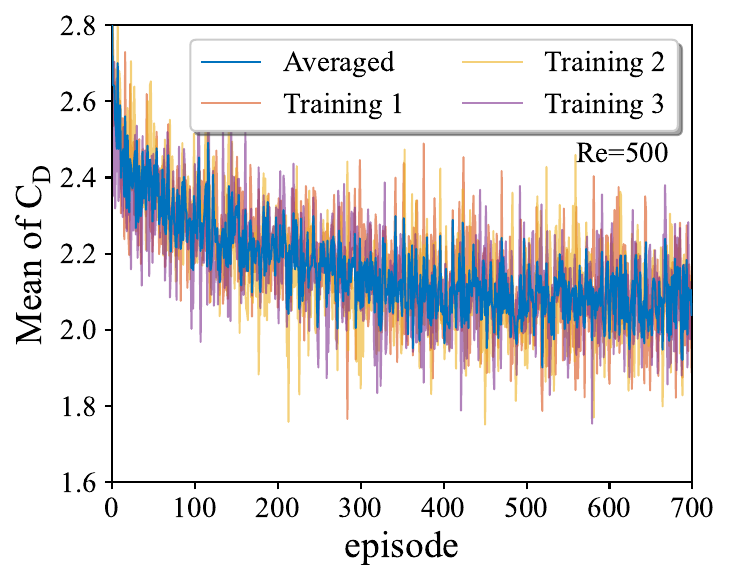}
        \put(-5,76){\fig{a}}
    \end{overpic}
    \hspace{2em}
    \begin{overpic}[width=0.45\textwidth]{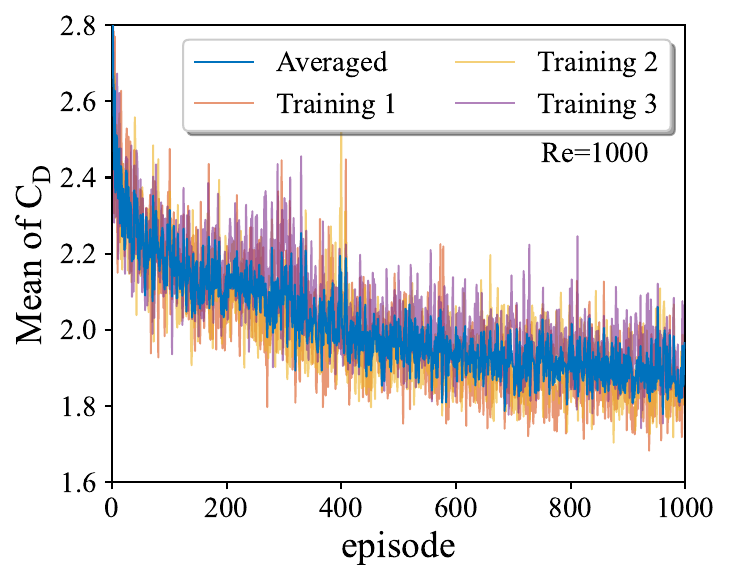}
        \put(-5,76){\fig{d}}
    \end{overpic} \\
    \begin{overpic}[width=0.45\textwidth]{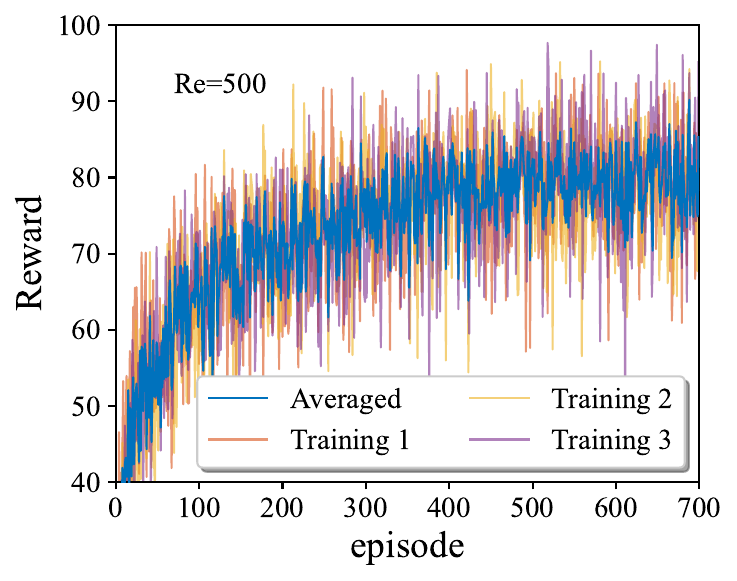}
        \put(-5,76){\fig{b}}
    \end{overpic}
    \hspace{2em}
    \begin{overpic}[width=0.45\textwidth]{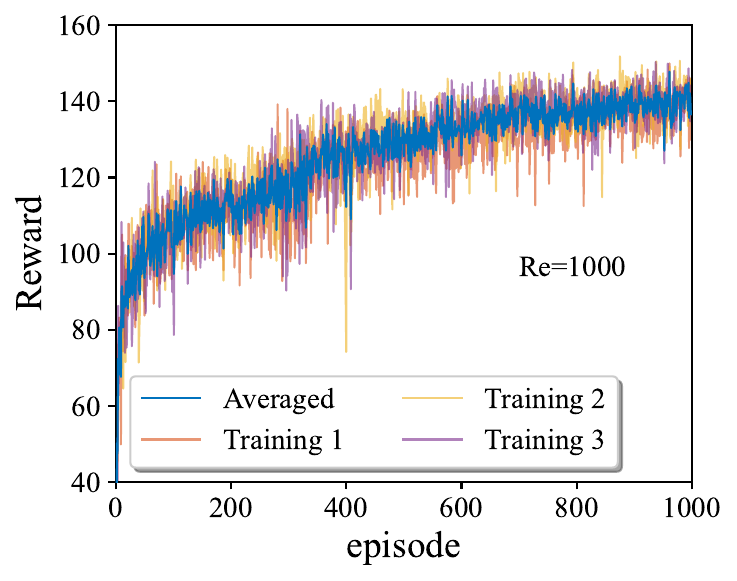}
        \put(-5,76){\fig{e}}
    \end{overpic} \\
    \begin{overpic}[width=0.45\textwidth]{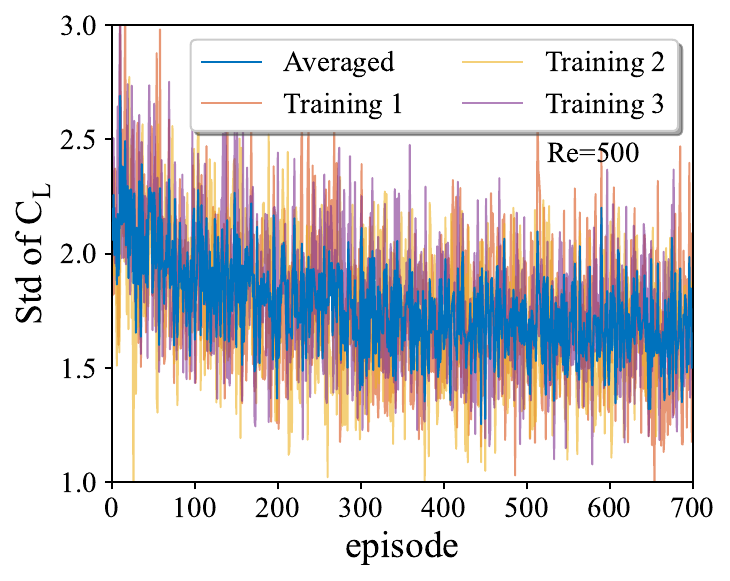}
        \put(-5,76){\fig{c}}
    \end{overpic} 
    \hspace{2em}
    \begin{overpic}[width=0.45\textwidth]{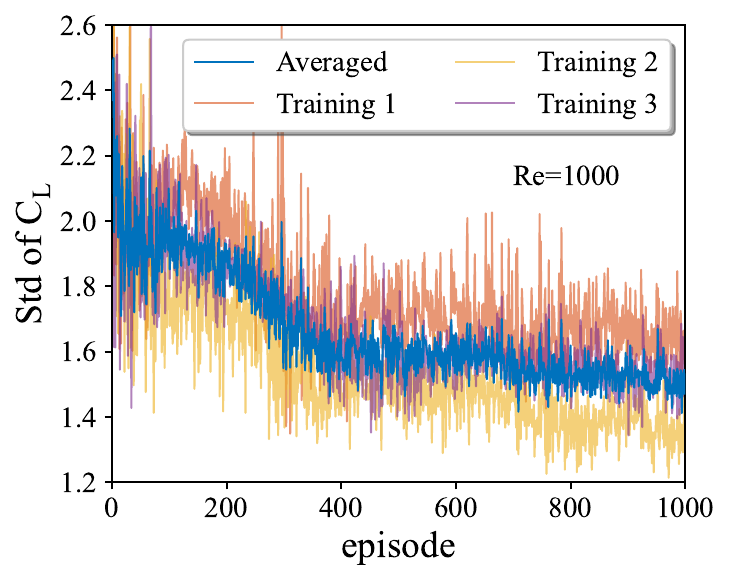}
        \put(-5,76){\fig{f}}
    \end{overpic}
\caption{Evolution of mean $C_D$ and reward curves in the learning process at $Re = 500$ [corresponding to \fig{a} and \fig{b}], $Re = 1000$ [corresponding to \fig{c} and \fig{d}]. The averaged curves labeled in blue refer to the moving average of the mean of all three training curves in each subgraph. All results are obtained under scheme \layout{3}{150}.} \label{fig:different_Re}
\end{figure}

Figure \ref{fig:different_Re} shows the evolution of the mean $C_D$, reward value, and $C_L$ obtained from three repeated training processes at Reynolds numbers of $Re$ = 500 and 1000. After approximately 400 episodes at $Re$ = 500 and 600 episodes at $Re$ = 1000, the $C_D$ approached convergence, demonstrating that a stable control strategy was achieved. Meanwhile, the reward curves gradually increased with each episode, with the std of $C_L$ declining and stabilizing. As described in Equation \ref{eq:reward_function}, $C_D$ and $C_L$ are first-order terms, where lift has a weight of 1, and drag has a weight of 0.1. For the deep reinforcement learning agent, this implies that the reduction of drag has a higher reward. When it is reduced to its maximum value (2.1 at $Re$ = 500 and 1.9 at $Re$ = 1000, respectively), inhibiting lift becomes the only viable option. However, as the Reynolds number increases, the learning requires more episodes to converge, and the agents need more trial-and-error steps to comprehend the nonlinear relationships inherent in the dynamic system.

An interesting phenomenon is observed when comparing the final performance of drag reduction across different Reynolds numbers. As the Reynolds number increases, the drag reduction effect improves. This contradicts our intuition, as we would normally expect the flow field to become more complex at high Reynolds numbers, with increased turbulence and vortex shedding, making it difficult for the DRL agent to learn an effective strategy for flow control. However, this is not the case. The main reason lies in the drag force component \citep{achenbachDistributionLocalPressure1968}. The overall drag $F_d$ on the circular cylinder submerged in a Newtonian fluid can be calculated by
\begin{gather}
{F_d} = \underbrace {\oint p  \cdot \cos (\theta ) \cdot {\text{d}}A}_{{\text{Pressure drag}}} + \underbrace {\oint {{\tau _{\text{w}}}}  \cdot \sin (\theta ) \cdot {\text{d}}A}_{{\text{Skin friction drag}}}, \\
\tau_w=\mu \cdot\left(\frac{\partial v_t}{\partial n}\right)_{\text {Surface}},
\end{gather}
where $p$ and $\tau _{\text{w}}$ are the normal stress and shear stress on the cylinder surface, respectively. $v_t$ is the velocity along the cylinder surface, and $n$ is the normal direction.

\begin{figure} \centering
    \begin{overpic}[width=0.35\textwidth]{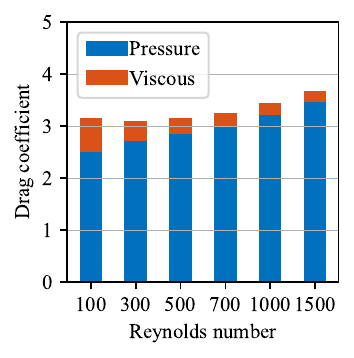}
        \put(0,90){\fig{a}}
    \end{overpic}
    \hspace{3em}
    \begin{overpic}[width=0.37\textwidth]{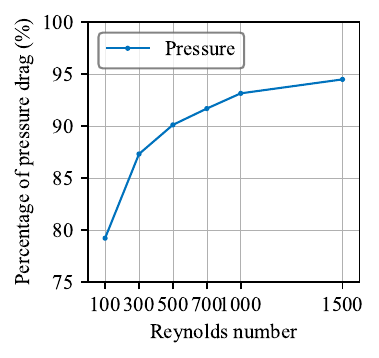}
        \put(-10,87){\fig{b}}
    \end{overpic} \\
\caption{Component analysis of \fig{a} Drag force caused by pressure and skin friction respectively at different Reynolds numbers, and \fig{b} the proportion of pressure drag in the total drag.}
\label{fig:component_analysis}
\end{figure}

Figure \ref{fig:component_analysis} shows the magnitudes of pressure drag, skin friction drag, and the proportion of pressure drag in total drag at different Reynolds numbers. It can be observed that, as the Reynolds number increases, the proportion of pressure drag monotonically increases, growing from 79.2\% at $Re=100$ to 94.5\% at $Re=1500$. Under an invariant aerodynamic shape and inflow velocity, the primary mechanism of active flow control was to suppress the shedding of vortices at the rear end of the cylinder \citep{wangFlowControlTechniques2018}. This indicates that the significant drag reduction effect was mainly attributed to the reduction of pressure drag caused by flow separation. As the Reynolds number increased, the proportion of pressure drag in the overall drag force increased, leading to a more pronounced drag reduction effect if the DRL controller effectively manipulates the flow separation.

\citet{tang2020robust} also proposed a compelling explanation. The flow around a cylinder can be decomposed into a superposition of steady baseflow and vortex-shedding components. The baseflow is numerically simulated using a symmetric boundary condition at the equatorial plane of the cylinder. The results showed that the drag force on the cylinder controlled by DRL was consistent with the drag force of the baseflow, which indicates that the drag reduction of active flow control using DRL mainly originates from vortex shedding, and the drag generated by vortex shedding is primarily attributable to the pressure drag component. Straightforwardly, under high Reynolds number conditions, the increased drag caused by the pressure component (both in absolute value and proportion) allows the DRL agent greater potential for flow control. When the DRL agent finds the optimal control rate, it leads to a decrease in the $C_D$.

\begin{figure} \centering
    \begin{overpic}[width=0.48\textwidth]{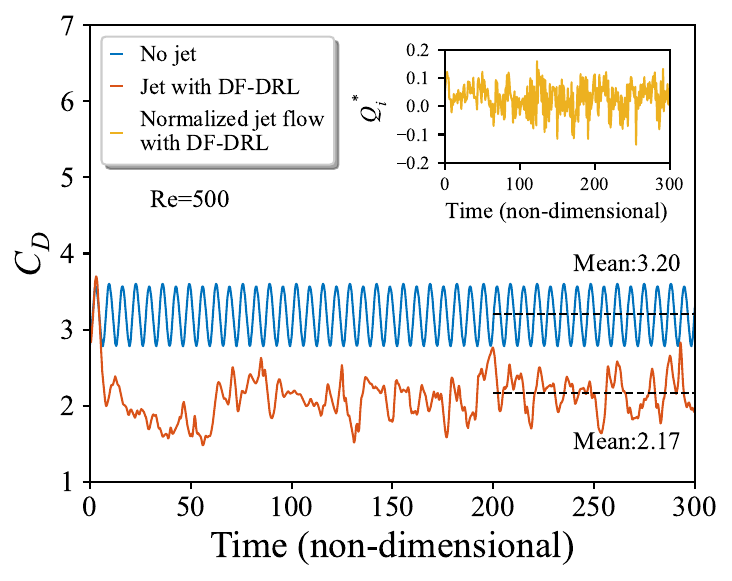}
        \put(0,75){\fig{a}}
    \end{overpic}
    \hspace*{\fill}
    \begin{overpic}[width=0.48\textwidth]{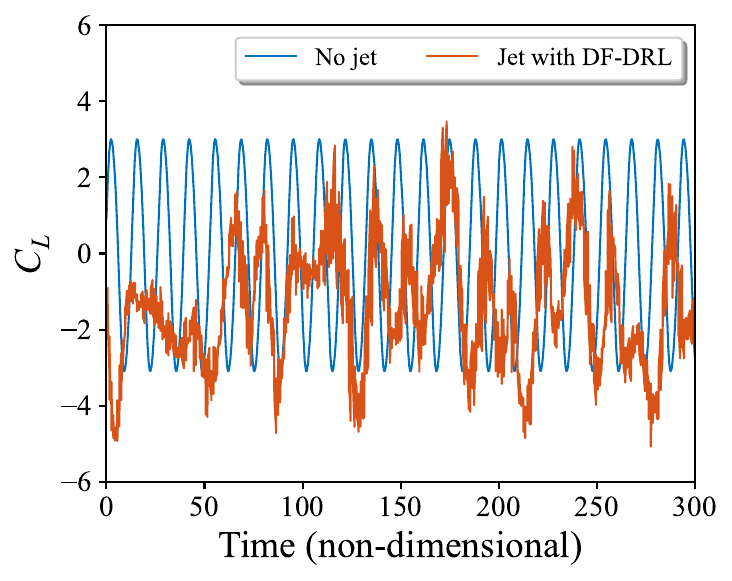}
        \put(0,75){\fig{b}}
    \end{overpic} \\
    \begin{overpic}[width=0.48\textwidth]{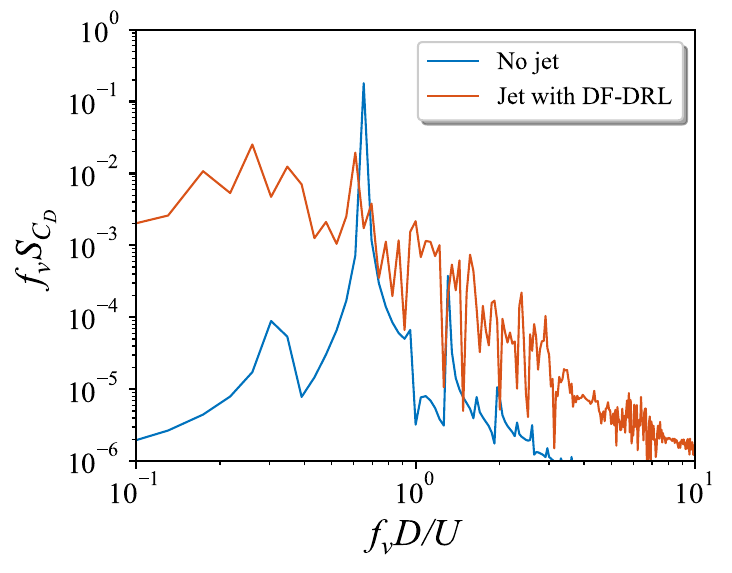}
        \put(0,75){\fig{c}}
    \end{overpic}
    \hspace*{\fill}
    \begin{overpic}[width=0.48\textwidth]{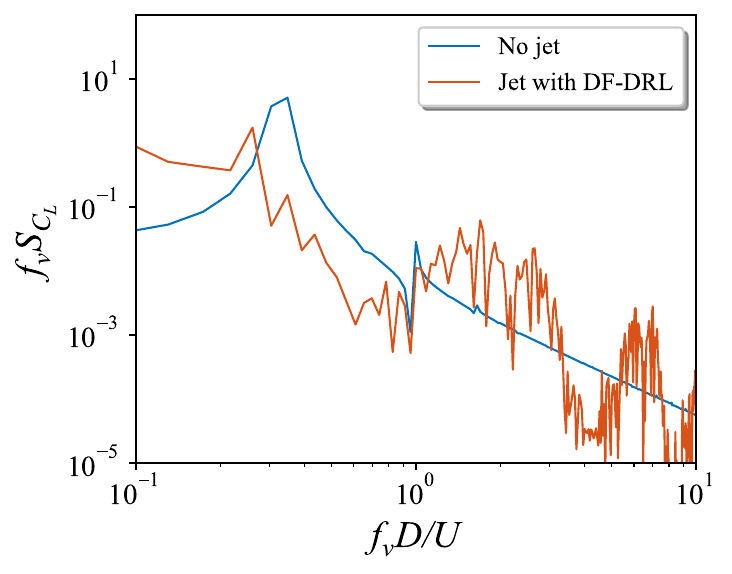}
        \put(0,75){\fig{d}}
    \end{overpic}
\caption{\fig{a} Evolution of $C_D$ for the cylinder without (no jet) and with (DF-DRL cylinder) active flow control at $Re = 500$, as well as the associated normalized flow rate of the jet flow;} \fig{b} Temporal variations in smoothed $C_L$ for the cylinder without (no jet) and with (DRL cylinder) active flow control in $Re = 500$; \fig{c} power spectral density (PSD) of the $C_D$ during the period of non-dimensional time ranging from 200 to 300; \fig{d} PSD of the $C_L$ during the period of non-dimensional time ranging from 200 to 300. \label{fig:500_cdcl}
\end{figure}

The results of the dynamic feature-based soft actor-critic (SAC) algorithm are presented in Fig. \ref{fig:500_cdcl}. The entire training process is parallelized across five environments provided by DRLinFluids. The algorithm successfully learned to perform active flow control, resulting in a continuous reduction of drag and suppression of lift. In the absence of actuation, the $C_D$ oscillates periodically around a mean value, as shown in Fig. \ref{fig:500_cdcl}\fig{a}. The mean value of $C_D$ is 3.20, with a std of 0.283 for the $C_D$ and 2.17 for the $C_L$. With DF-DRL-based active flow control, the mean $C_D$ is reduced to 2.17, corresponding to a drag reduction of approximately $32.2\%$. Furthermore, the fluctuation of $C_D$ is suppressed, as indicated by the reduced std value of 0.252. Meanwhile, the std value of $C_L$ is decreased slightly to 1.61.

Power spectrum analyses are conducted to compare the $C_D$ and $C_L$ of the cylinder with and without active flow control, and the results are presented in Fig. \ref{fig:500_cdcl}\fig{c} and \fig{d}. The power spectrum curves for both $C_D$ and $C_L$ of the plain cylinder exhibit a distinct peak. This indicates a series of distinct vortex shedding at this frequency, contributing to the majority of the energy required for the mean drag and the fluctuation. By contrast, the peaks disappear in the power spectrum curves of $C_D$ and $C_L$ of the cylinder with DF-DRL-based active flow control.

\begin{figure} \centering
    \begin{overpic}[width=0.48\textwidth]{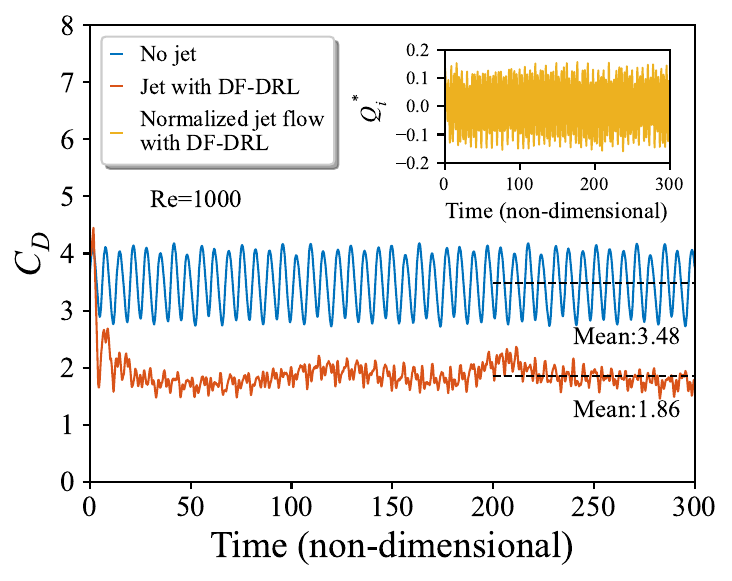}
        \put(0,75){\fig{a}}
    \end{overpic}
    \hspace*{\fill}
    \begin{overpic}[width=0.48\textwidth]{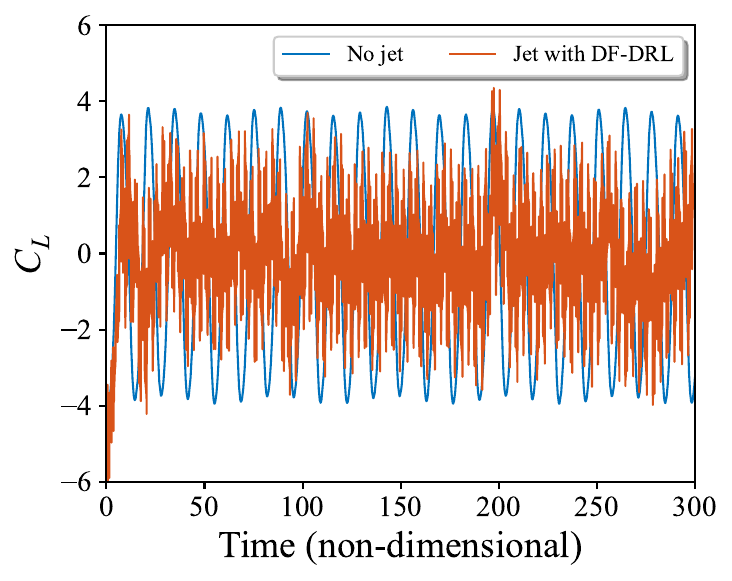}
        \put(0,75){\fig{b}}
    \end{overpic} \\
    \begin{overpic}[width=0.48\textwidth]{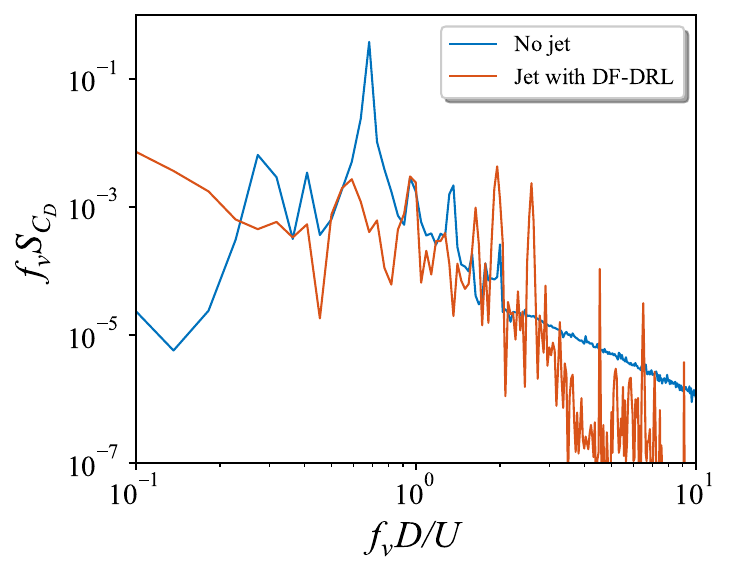}
        \put(0,75){\fig{c}}
    \end{overpic}
    \hspace*{\fill}
    \begin{overpic}[width=0.48\textwidth]{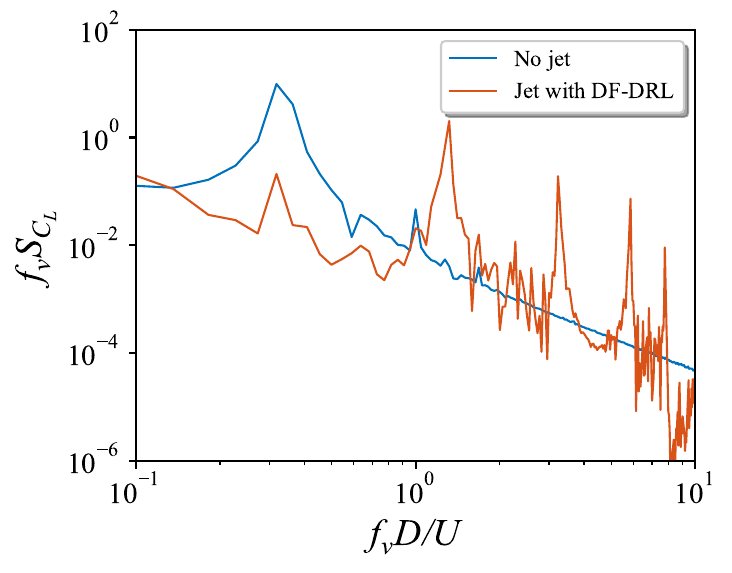}
        \put(0,75){\fig{d}}
    \end{overpic}
\caption{\fig{a} Evolution of $C_D$ for the cylinder without (no jet) and with (DF-DRL cylinder) active flow control at $Re = 1000$, as well as the associated normalized flow rate of the jet flow; \fig{b} Temporal variations in smoothed $C_L$ for the cylinder without (no jet) and with (DF-DRL cylinder) active flow control at $Re = 1000$; \fig{c} and \fig{d} are the PSD of $C_D$ and $C_L$ respectively during the period of non-dimensional time ranging from 200 to 300.} \label{fig:1000_cdcl}
\end{figure}

The results presented in Fig. \ref{fig:1000_cdcl} demonstrate that the turbulent conditions are relatively weak at a Reynolds number of 1000. Precisely, for the plain cylinder, the mean $C_D$ is measured to be 3.48 with a std of 0.455. In addition, the $C_L$ exhibits a std of 2.76. However, when active flow control is implemented, a significant reduction in the mean $C_D$ is achieved, resulting in a value of 1.86, corresponding to a drag reduction of approximately 46.55$\%$. Moreover, the std of $C_D$ decreases to 0.31, indicating a more consistent behavior of the cylinder under flow control. Notably, the std of $C_L$ is also markedly reduced to 1.61, which is highly desirable for suppressing the lift force and mitigating flow-induced instability of the cylinder.

The power spectrum analyses of $C_D$ and $C_L$ for the cylinder with and without active flow control are presented in Fig. \ref{fig:1000_cdcl}\fig{c} and \fig{d}. The power spectrum curves for both $C_D$ and $C_L$ of the plain cylinder show an obvious peak, indicating the presence of a regular vortex shedding significant energy. In contrast, when active flow control is implemented, the peak in the power spectrum curves of $C_D$ and $C_L$ for the cylinder is eliminated, indicating that the jet actuation has completely disrupted the regular vortex shedding.

\begin{figure} \centering
    \begin{overpic}[width=0.8\textwidth]{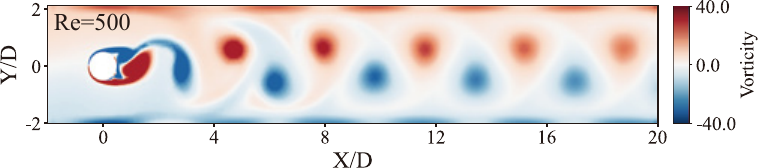}
        \put(-1,20){\fig{a}}
    \end{overpic} \\
    \begin{overpic}[width=0.8\textwidth]{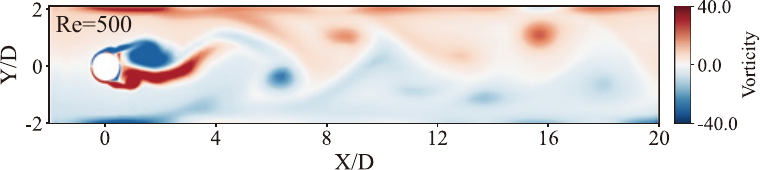}
        \put(-1,20){\fig{b}}
    \end{overpic} \\
    \begin{overpic}[width=0.8\textwidth]{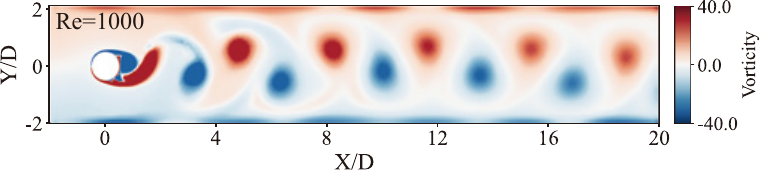}
        \put(-1,20){\fig{c}}
    \end{overpic} \\
    \begin{overpic}[width=0.8\textwidth]{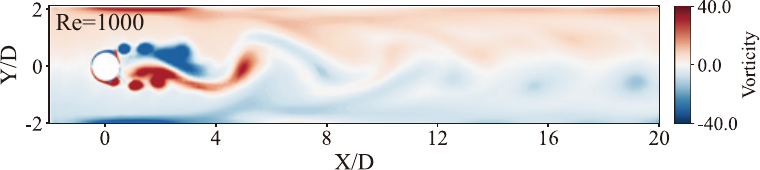}
        \put(-1,20){\fig{d}}
    \end{overpic}
\caption{Typical snapshots of the vorticity channel around the cylinder at $Re = 500$ corresponding to the flow without \fig{a} and with \fig{b} control and $Re = 1000$ corresponding to the flow without \fig{c} and with \fig{d} control.} \label{fig:500_1000shunshiwoliangtu}
\end{figure}

Figure \ref{fig:500_1000shunshiwoliangtu} displays the instantaneous flow field around a circular cylinder, with and without active control. The impact of controlled jet flow on reducing the aerodynamic force acting on the cylinder is explained in terms of the flow pattern. In Fig. \ref{fig:500_1000shunshiwoliangtu}\fig{a} and \fig{c}, which represent conditions for $Re = 500$ and 1000, respectively, a vortex shedding pattern is observed for the plain cylinder, as expected. This alternate vortex shedding pattern directly contributes to fluctuations in both $C_D$ and $C_L$, as demonstrated in Fig. \ref{fig:500_cdcl} and \ref{fig:1000_cdcl}.

Figure \ref{fig:500_1000shunshiwoliangtu}\fig{b} and \fig{d} illustrate the impact of fluctuating actuation on the vortex shedding pattern. The alternate vortex shedding is suppressed, reducing fluctuations of both $C_D$ and $C_L$. Meanwhile, an elongated recirculation bubble is formed in the near wake, associated with increased pressure and a reduction in drag force. The elongated wake implies a reduced curvature of the shear layer, corresponding to increased pressure at the rearward cylinder side. As a result, the cylinder with active flow control experiences less drag.

\begin{figure} \centering
    \begin{overpic}[width=0.8\textwidth]{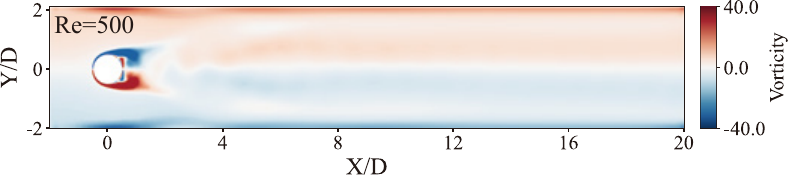}
        \put(-1,20){\fig{a}}
    \end{overpic} \\
    \begin{overpic}[width=0.8\textwidth]{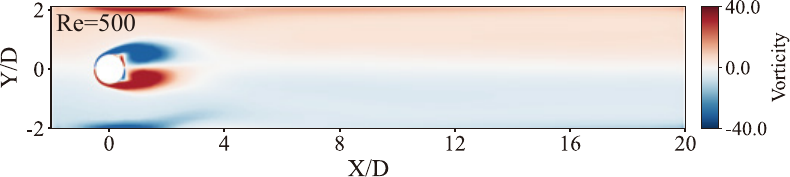}
        \put(-1,20){\fig{b}}
    \end{overpic} \\
    \begin{overpic}[width=0.8\textwidth]{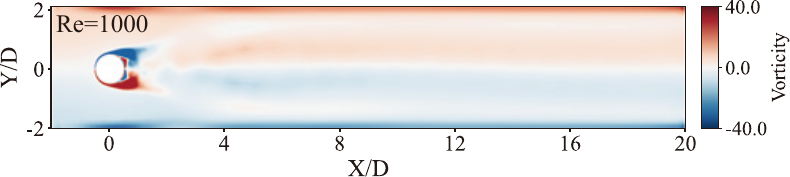}
        \put(-1,20){\fig{c}}
    \end{overpic} \\
    \begin{overpic}[width=0.8\textwidth]{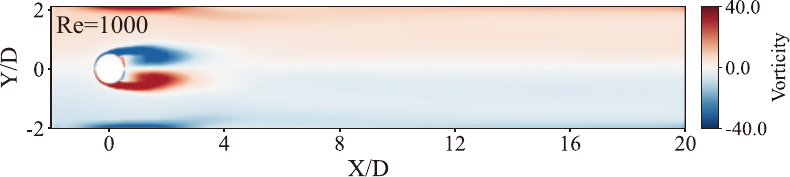}
        \put(-1,20){\fig{d}}
    \end{overpic}
\caption{Mean vorticity contours around the cylinder at $Re = 500$ corresponding to the flow without \fig{a} and with \fig{b} control and $Re = 1000$ corresponding to the flow without \fig{c} and with \fig{d} control.} \label{fig:500_1000pingjunwoliangtu}
\end{figure}

Figure \ref{fig:500_1000pingjunwoliangtu} depicts the mean vorticity contours around a cylinder, with and without active control. The increase in the recirculation zone is obvious at $Re$ = 500 and 1000 and illustrates the effective control strategy learned by the DRL agent. The results demonstrate that well-trained DRL agents, which utilize a single surface pressure sensor's temporal information as the state, can achieve efficient control even under flow conditions with strong nonlinearity and various Reynolds numbers.

It is worth noting that learning a DRL-based control law for the AFC task presents a significant challenge in utilizing a single surface pressure sensor as the state in weak turbulent conditions. However, these results demonstrate the efficacy of DF-DRL-based active flow control of a circular cylinder with sparse surface pressure sensing and offer a promising avenue for reducing drag and enhancing AFC performance in fluid dynamics systems.

\subsubsection{Case 2: 3D turbulent flow around a circular cylinder}
\begin{figure} \centering
\begin{overpic}[width=0.7\textwidth]{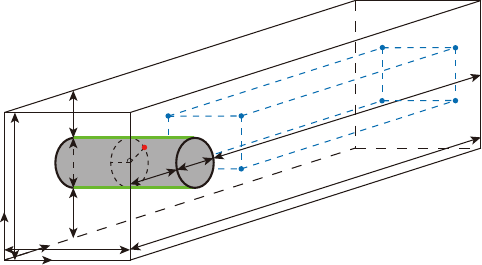}
    \put(4,14){\rot{90}{21D}}
    \put(16,8){\rot{90}{10D}}
    \put(16,19){\rot{90}{1D}}
    \put(16,28){\rot{90}{10D}}
    \put(28.5,19){\rot{13}{10D}}
    \put(38,22){\rot{13}{1D}}
    \put(60,28.5){\rot{13}{25.5D}}
    \put(18,4){4D}
    \put(60,15){\rot{13}{36.5D}}
    \put(25,22){$\theta$}
    \put(9,5){\bm{$x$}}
    \put(-2,10){\bm{$y$}}
    \put(10,-2){\bm{$z$}}
\end{overpic}
\caption{Schematics of the computational domain for three-dimensional turbulent flow around a cylinder. The red dot represents the locations of a surface pressure sensor, with 0 degrees corresponding to the most front of the cylinder and theta equal to 135 degrees. The green bar-shaped regions on either side of the cylinder represent the jet actuators. The blue boxes depict sensor clusters deployed in the wake region. The wake sensors are deployed at equal intervals along the $x$, $y$, and $z$ direction, forming an $8 \times 3 \times 3$ grid configuration.} \label{fig:3d_flow_over_cylinder}
\end{figure}

Further consideration is given to a more complex flow scenario that closely resembles real-world conditions. Turbulent flow around a 3D cylinder with a circular cross–section at \Rey =10000 is studied in this section. The entire computational domain was discretized using polyhedral meshes of varying sizes. The mesh is directly generated from the three-dimensional simulation region to accurately simulate the physics and minimize numerical diffusion, consisting of a total of 1.33 million grids and 15 layers of boundary mesh on the surface of the cylinder. The dimensions of the computational domain are set following the study conducted by \citet{navroseFreeVibrationsCylinder2013}. The distances from the inlet (upstream) and outlet (downstream) to the surface of the cylinder are set at 10D and 25.5D, respectively. Additionally, the distances from the top and bottom surfaces to the cylinder surface are both 10D. The cylinder has a spanwise distance of 4D, which has been depicted as Fig. \ref{fig:3d_flow_over_cylinder}. 

The velocity on the surface of the cylinder is subject to a no-slip condition, ensuring that the fluid adheres to the cylinder's surface. At the upstream boundary, free-stream values are assigned to the velocity. The stress vector is set to zero at the downstream boundary. On the remaining boundaries, the normal component of the velocity and the tangential component of the stress vector are prescribed a zero value in both directions. Throughout the time-marching solution process, the position of the cylinder, its velocity, and the boundary conditions are updated at each nonlinear iteration to capture the evolving flow dynamics accurately. The results of mesh convergence are listed in Table \ref{tab:3d_flow_validation}.
\begin{table}
    \centering
    \caption{Comparison of integral flow quantities in the flow past a 3D circular cylinder at \Rey = 10000. $C_d$ is drag coefficient, $C_L$ is r.m.s. value of lift coefficient, and $S_t$ is the Strouhal number.} \label{tab:3d_flow_validation}
        \begin{tabularx}{\textwidth}{LCCC}
            \toprule
              & Mean of $C_d$ & r.m.s. of $C_L$ & $S_t$ \\
            \midrule
            \citet{gopalkrishnanVortexinducedForcesOscillating1993}         & 1.186 & 0.384 & 0.193 \\
            \citet{bishopLiftDragForces1997}      & -     & 0.463 & 0.201 \\
            \citet{norbergFluctuatingLiftCircular2003}              & -     & 0.394 & 0.202 \\
            \citet{dongCombinedDirectNumerical2006} & 1.143 & 0.448 & 0.203 \\
            \citet{fanReinforcementLearningBluff2020} & 1.192 & 0.482 & 0.204 \\
            Present & 1.151 & 0.431 & 0.200 \\
            \bottomrule
        \end{tabularx}
\end{table}

\begin{figure} \centering
    \begin{overpic}[width=0.48\textwidth]{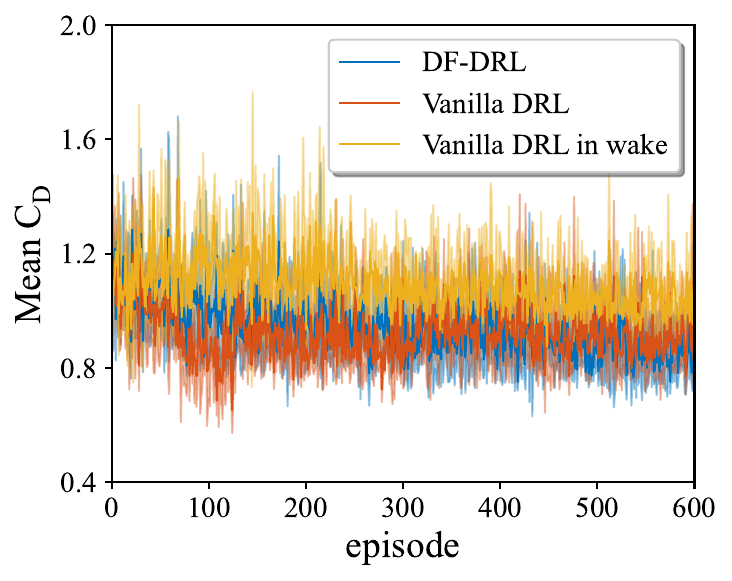}
        \put(0,75){\fig{a}}
    \end{overpic}
    \hspace*{\fill}
    \begin{overpic}[width=0.48\textwidth]{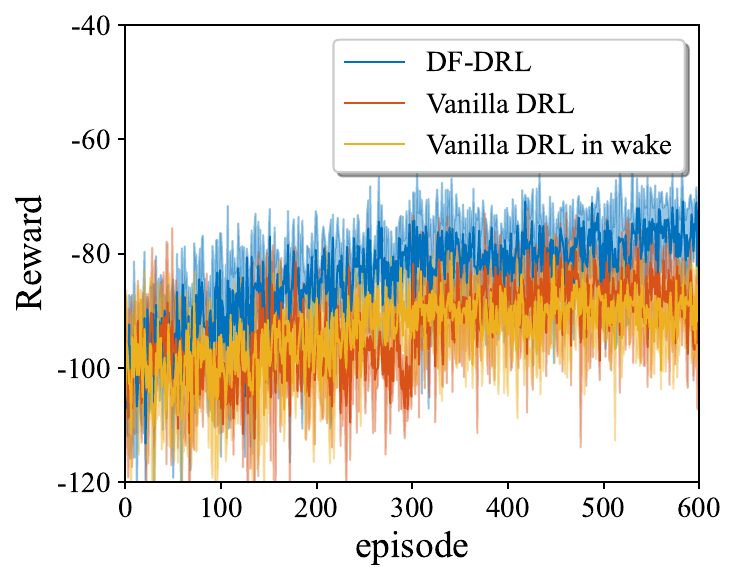}
        \put(0,75){\fig{b}}
    \end{overpic}
    \\
    \begin{overpic}[width=0.48\textwidth]{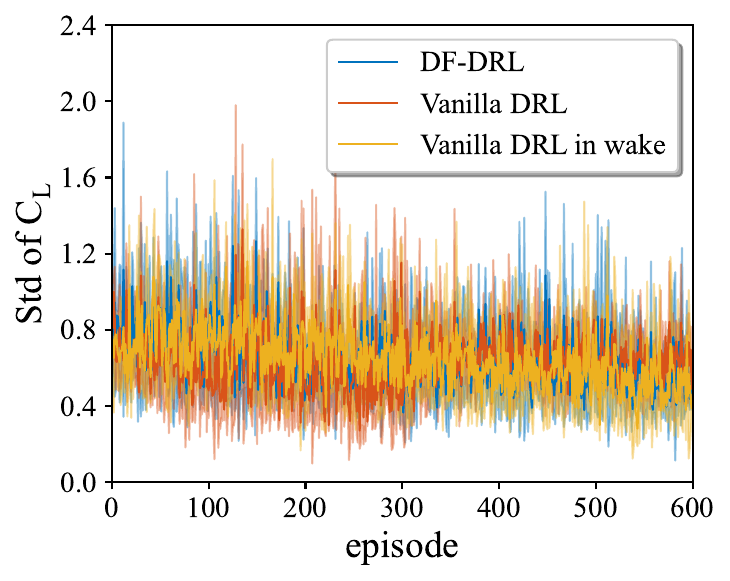}
        \put(0,75){\fig{c}}
    \end{overpic}
    \caption{Evolution of \fig{a} mean drag reduction, \fig{b} reward, and \fig{c} std lift coefficient of vanilla DRL and DF-DRL method with surface pressure sensing, and vanilla DRL method with wake sensing, respectively. A detailed description of the layout schemes for the two types of sensors is provided in Fig. \ref{fig:3d_flow_over_cylinder}.} \label{fig:lc_re10000}
\end{figure}

Subsequently, we conducted active flow control experiments on three different configurations for three-dimensional flow around a cylinder. These configurations include: (Scheme A) flow control using a single surface pressure sensor as the state input for a DF-DRL based actuator; (Scheme B) flow control using a single surface pressure sensor as the state input for a vanilla DRL-based actuator; and (Scheme C) flow control using wake filed velocity sensors as the state input for a vanilla DRL-based actuator. The results of the learning curves during the training stage are shown in Fig. \ref{fig:lc_re10000}.

Figure \ref{fig:lc_re10000}\fig{a} shows the evolution of the mean drag coefficient along with the increase of DRL training episodes with and without dynamic feature lifting. The results indicate that both the method and the vanilla DRL method achieve significant drag-reduction effects. Compared to the case without control, the mean drag was reduced to 0.822, 0.867, and 1.01 for schemes A, B, and C, respectively. However, some significant differences can be observed from the reward learning curves as described in Fig. \ref{fig:lc_re10000}\fig{b}. The reward curve based on the DF-DRL method shows a steadily increasing trend throughout the entire episode, eventually reaching around -76. On the other hand, vanilla DRL, especially with a surface pressure sensor, exhibits a noticeable instability between episodes 50 and 300, characterized by large fluctuations, followed by a relatively low value near 320. Another factor contributing to the difference in reward values is the standard deviation of the lift coefficient, as shown in Fig. \ref{fig:lc_re10000}\fig{c}. The scheme B, and C increase lift fluctuations to 0.626 and 0.526, while scheme A with the DF-DRL method is more effective in suppressing lift fluctuations, resulting in 0.394.
\begin{figure} \centering
    \begin{overpic}[width=0.48\textwidth]{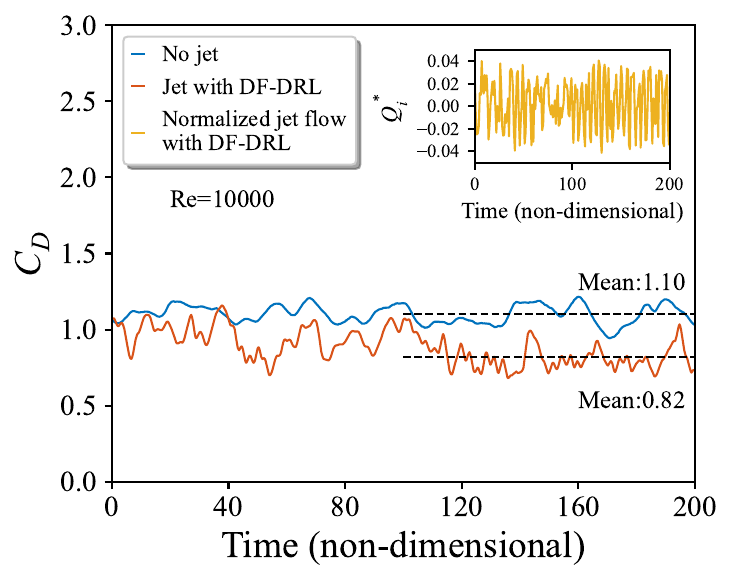}
        \put(0,75){\fig{a}}
    \end{overpic}
    \hspace*{\fill}
    \begin{overpic}[width=0.48\textwidth]{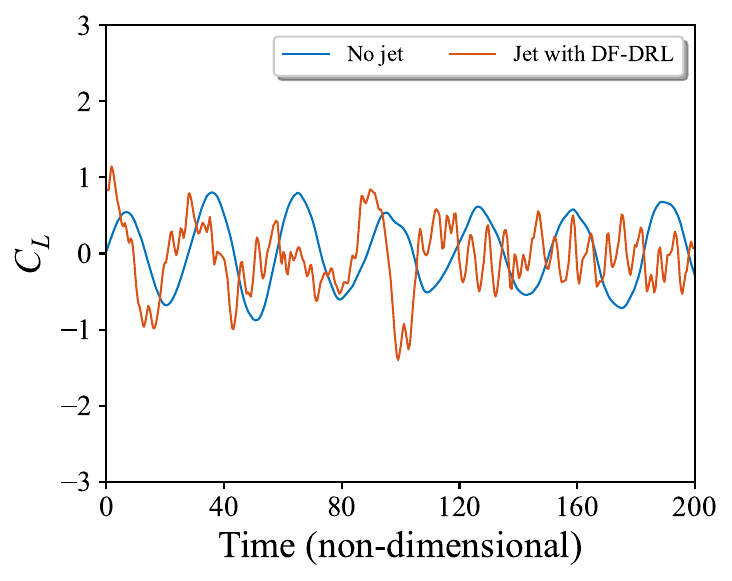}
        \put(0,75){\fig{b}}
    \end{overpic} \\
    \begin{overpic}[width=0.48\textwidth]{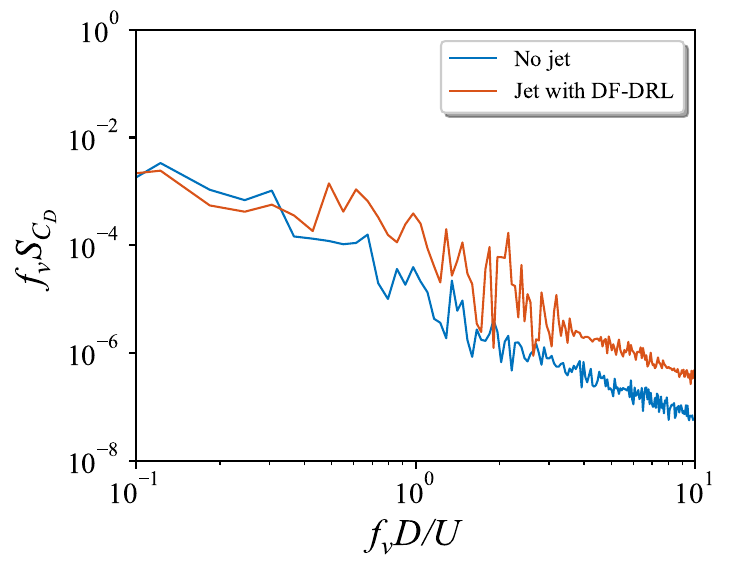}
        \put(0,75){\fig{c}}
    \end{overpic}
    \hspace*{\fill}
    \begin{overpic}[width=0.48\textwidth]{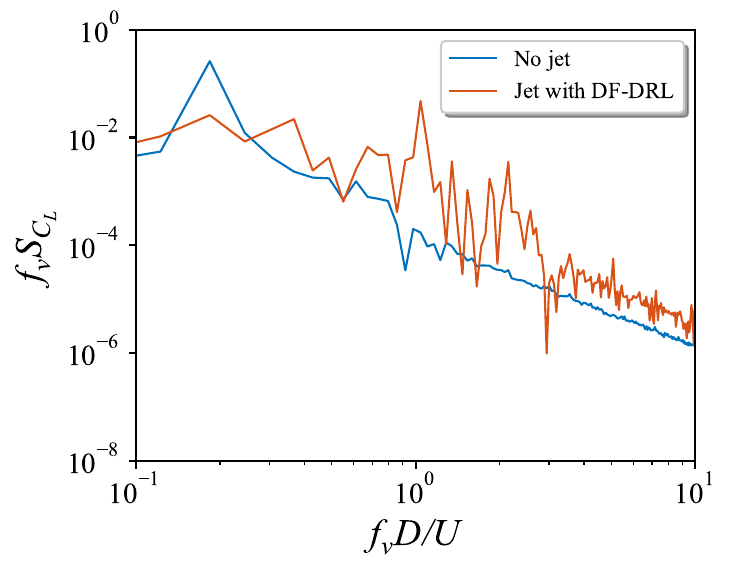}
        \put(0,75){\fig{d}}
    \end{overpic}
\caption{\fig{a} Evolution of $C_D$ for the cylinder without (no jet) and with (DF-DRL cylinder) active flow control at \Rey = 10000, {as well as the associated normalized flow rate of the jet flow $Q_i^*$ (yellow line), which is defined in Eq. \eqref{flowrate}}; \fig{b} Temporal variations in smoothed $C_L$ for the cylinder without (no jet) and with (DF-DRL cylinder) active flow control at \Rey = 10000; \fig{c} and \fig{d} are the PSD of $C_D$ and $C_L$ respectively during the period of non-dimensional time ranging from 100 to 200.} \label{fig:ts_DF-DRL_re10000}
\end{figure}

{To further clarify the minor differences in drag reduction observed between the scheme A and scheme B, additional experiments are conducted under different reward function settings, notably those excluding the lift coefficient penalty in Vanilla DRL training. The outcomes of these experiments align with the findings of \citet{Rabault2019Artificial}, wherein the omission of the penalty term leads to the adoption of a ``cheating” control strategy by the agent. This strategy involves a significant shift towards unilateral continuous jet actuation, resulting in a decrease in the lift coefficient to 0.62, marking a reduction of 43.6\%. Concurrently, an increase in the r.m.s. of lift from 1.77 to 6.08 is observed, representing an increase of 243.5\%. The implementation of such a control strategy, while beneficial for reducing drag, induces fluctuations in lift that could be deemed unacceptable for certain practical applications. This highlights the critical need for a well-balanced reward function in DRL applications, ensuring that the control strategies employed do not compromise the structural stability and performance in practical scenarios.}

After a training phase consisting of 600 episodes, the DF-DRL agent (scheme A) is further utilized for testing purposes. The results, as shown in Fig. \ref{fig:ts_DF-DRL_re10000}, reveal a decrease in the drag coefficient from its initial value of 1.151 to 0.822, achieving a drag reduction of 28.6\%, meanwhile, the r.m.s. of lift coefficient decreases from 0.431 to 0.394. Figure \ref{fig:ts_DF-DRL_re10000}\fig{c} and \fig{d} demonstrate that DRL-based AFC primarily suppresses drag and lift forces by inhibiting energy in the relatively low-frequency region, which is the dominant of the energy associated with drag and lift forces. {It also should be emphasized that the actuator's behavior significantly surpasses the basic dichotomy of either continuous blowing or suction (at a constant value) and diverges from the typical operation of a bang-bang controller at its maximum output. Instead, the actuator matures into an adaptive, real-time controller, adept at dynamically responding to the environment's changing conditions.}

\begin{figure} \centering
    \hspace*{\fill}
    \begin{overpic}[width=0.45\textwidth]{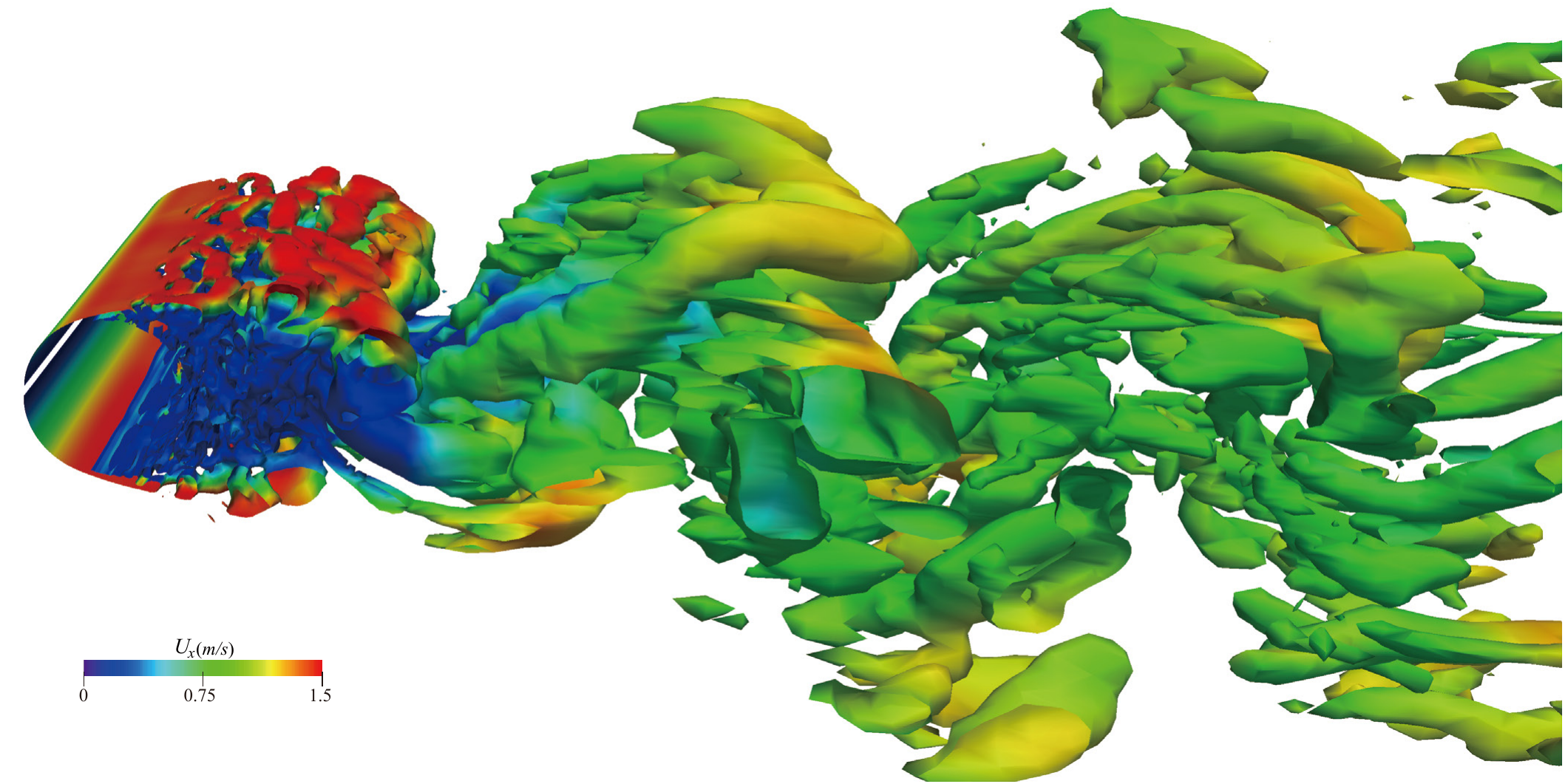}
        \put(-3,50){\fig{a}}
    \end{overpic}
    \hspace*{\fill}
    \begin{overpic}[width=0.45\textwidth]{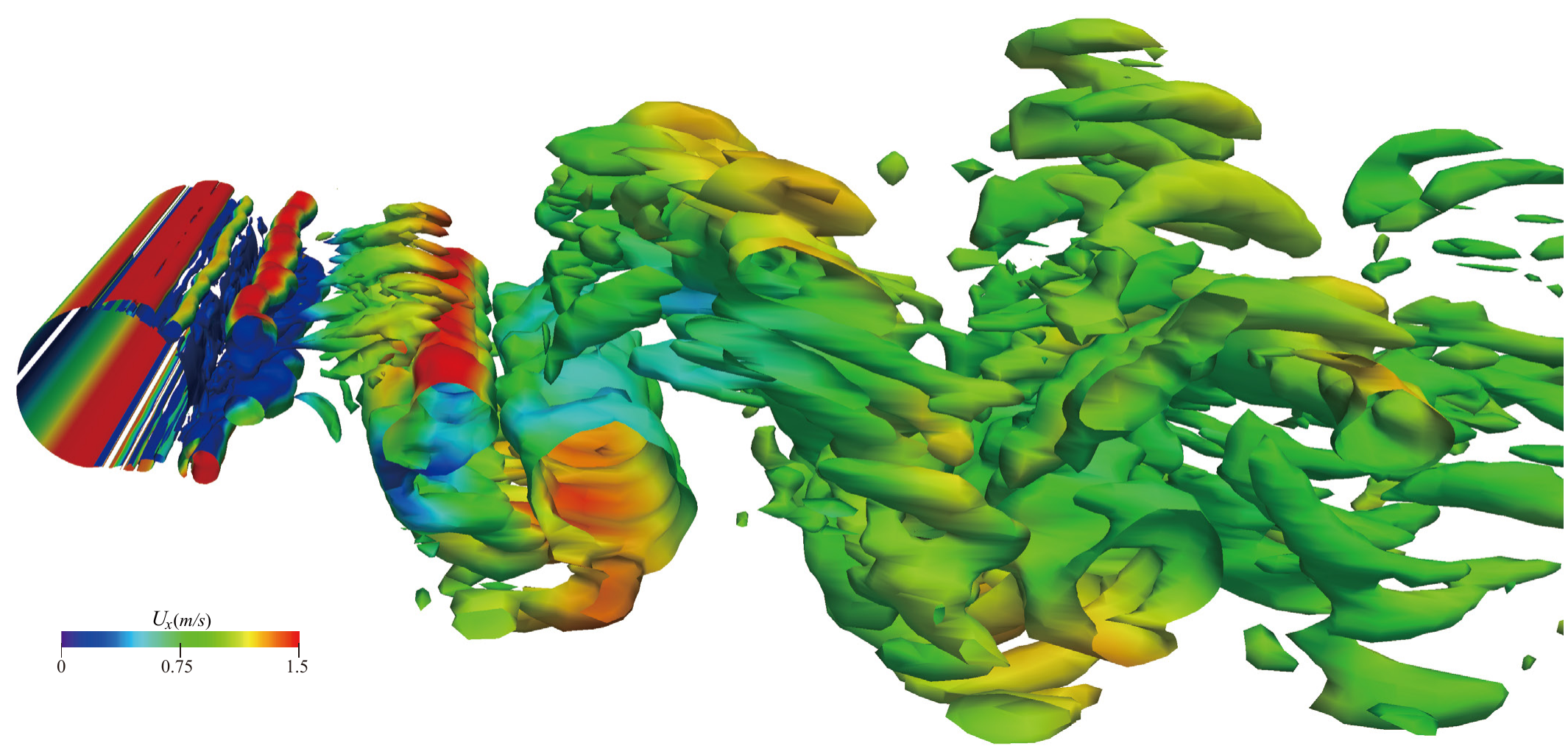}
        \put(-3,50){\fig{b}}
    \end{overpic}
    \hspace*{\fill}
    \\ \vspace{1em}
    \hspace*{\fill}
    \begin{overpic}[width=0.45\textwidth]{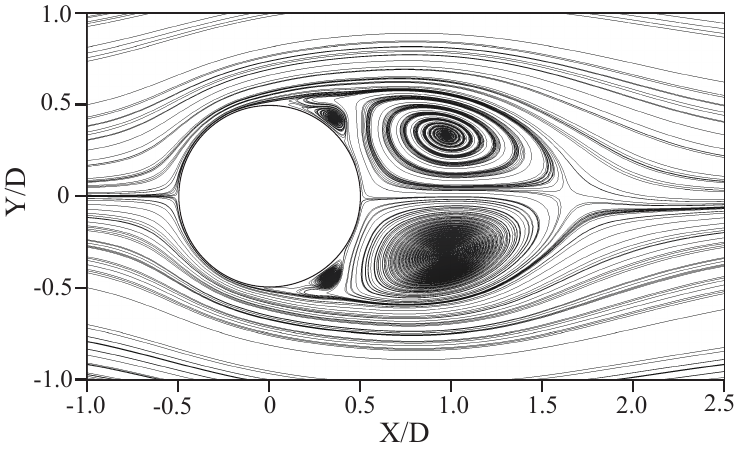}
        \put(-3,58){\fig{c}}
    \end{overpic}
    \hspace*{\fill}
    \begin{overpic}[width=0.45\textwidth]{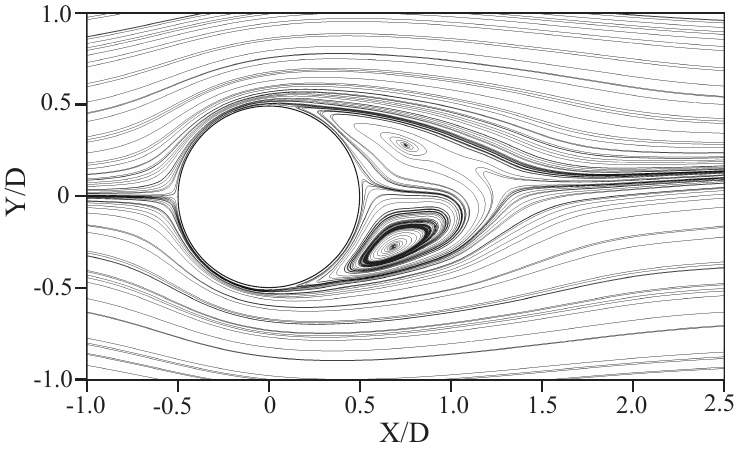}
        \put(-3,58){\fig{d}}
    \end{overpic}
    \hspace*{\fill}
\caption{Turbulent flow past a circular cylinder at Re = 10000. Instantaneous vortical structures close to the cylinder surface colored by the streamwise velocity \fig{a} without and \fig{b} with DF-DRL based control, and streamlines of mean flow \fig{c} without and \fig{d} with DF-DRL based control.} \label{fig:q_and_mean_streamline}
\end{figure}

From the perspective of three-dimensional vortex structures in the wake field, as shown in Fig. \ref{fig:q_and_mean_streamline}\fig{a} and \fig{b}, it is evident that in the case without any jet injection, small and fragmented vortices are generated in the wake region near the cylinder. As the flow moves away from the cylinder, the smaller-scale vortex structures gradually dissipate, making way for larger-scale vortices. However, when the DF-DRL based actuator is employed for control, a more regular elongated vortex structure is formed as a result of the blowing and suction of air on both sides, combined with the incoming flow.

This elongated vortex structure, resembling a strip created by the actuator, plays a role in mitigating the generation of fragmented vortices to some extent. As a result, the mid to far wake region exhibits a more regular and alternating pattern of elongated vortices.

The utilization of the DF-DRL based actuator introduces a controlled airflow that has a significant impact on the wake flow characteristics. The formation of the elongated vortex structure with more regularity and coherence indicates an improved flow control capability. This controlled vortex structure has the potential to enhance aerodynamic performance, reduce drag, or achieve other desired flow control objectives in various engineering applications.

The dimensions of the wake recirculation zone, specifically its width, and length, directly influence the base pressure behind a two-dimensional bluff body \citep{heDragReductionSquare2014,royModulationRecirculationRegion2019}. As shown in Fig. \ref{fig:q_and_mean_streamline}\fig{c} and \fig{d}, the wake region of the cylinder, controlled by the DF-DRL based actuator, exhibits the characteristics compared to the uncontrolled cylinder, including (1) the disappearance of secondary vortices, (2) a reduction of 37.1\% in the maximum width of the wake, and (3) a slight decrease of 15\% in the length of the wake. The first two factors lead to a significant decrease in drag, which outweighs the adverse effect of a shorter recirculation zone on drag. This overall effect manifests as a reduction in drag, resulting in a decrease of approximately 28.6\%.

\section{Conclusions and outlook}
This study presents significant progress towards practical active flow control with surface pressure sensors located on a circular cylinder as the sole input for a DRL agent. This approach can potentially advance deep reinforcement learning for real-world applications, such as drag and lift reduction for vehicle and high-rise buildings. The main results of this study are summarized as follows.

Firstly, a novel deep reinforcement learning method called Dynamic Feature-Based Deep Reinforcement Learning (DF-DRL) is introduced. Essentially, DF-DRL utilizes prior knowledge to extract one or several features of a nonlinear dynamic system, enabling it to estimate the complete state of the system to the fullest extent possible. This concept aligns with the ideas of pattern recognition and reduced order modeling. The DF-DRL model combines identification and control, which is not limited to the traditional DRL tasks that take the state at a certain moment as input. Instead, suitable dynamic feature states are selected and lifted to a higher dimensional vector based on the characteristics of different dynamic systems. Following this step, the vector is used as the state input to the agent. Results show that DF-DRL with a single surface pressure sensor can achieve the same drag reduction performance as the vanilla DRL method using 147 velocity sensors that fully sample the cylinder wake region.

Secondly, the study investigates the distribution of sensors needed for active flow control of cylinder wake. We conclude that in low to moderate Reynolds number scenarios, a single surface pressure sensor can achieve control results comparable to those obtained with 147 wake sensors under active flow control when DF-DRL is used. Additionally, we find that the reward value obtained with a single trailing edge sensor on the cylinder is higher than if the sensor is located at the leading edge, resulting in a lower mean $C_D$ and the std of $C_L$.

Thirdly, three different flow configurations were examined to verify the effectiveness and robustness of the proposed sensor configuration and DF-DRL method. Results show that the deep reinforcement learning agent utilizing a single surface pressure sensor is capable of controlling wake development behind the circular cylinder, even under more complex scenarios corresponding to higher Reynolds numbers or 3D turbulent inflow.

Sparse reduced-order modeling \citep{Brunton2016pnas,Loiseau2018jfm} is a highly popular research field in which selecting appropriate dynamic feature lifting methods for different fluid dynamic systems can enable more accurate estimation using fewer sensor data. Processing these features and using them as DRL states is a promising approach. In the present study, significant reductions in the $C_D$ of a cylinder are achieved through two distinct approaches. Specifically, the use of typical DRL resulted in a reduction of 6.4\% utilizing a 4 sensors layout scheme, while dynamic feature sensing with lifting yielded a reduction of 8\% compared to the benchmark performance under a low Reynolds number. Under a turbulent flow around a cylinder at high Reynolds numbers, there is a more significant reduction in drag coefficient, reaching as high as 28.6\% with a DF-DRL controller. The results of this investigation demonstrate that the $C_D$ of the dynamic feature sensing with lifting and DRL (DF-DRL) model is impressively lower than the vanilla model that relies solely on direct sensor feedback, highlighting the efficacy of this approach for improving aerodynamic performance. The DF-DRL method presents a promising approach to significantly reducing the number of required sensors while achieving optimal $C_D$ and $C_L$ reduction performance, which offers a promising pathway for taming complex fluid dynamics systems.

\section*{Acknowledgement}
This study is supported by the National Key R\&D Program of China (2021YFC3100702), National Natural Science Foundation of China (52278493, 52108451), Shenzhen Science and Technology Program (SGDX20210823103202018, GXWD20201230155427003-20200823230021001, KQTD20210811090112003), and Guangdong-Hong Kong-Macao Joint Laboratory for Data-Driven Fluid Mechanics and Engineering Applications (2020B1212030001).
This work is also supported 
by the National Science Foundation of China (NSFC) through grants 12172109 and 12172111, %
by Guangdong province, China, via the Natural Science and Engineering grant 2022A1515011492
and by the Shenzhen Research Foundation for Basic Research, China, through grant JCYJ20220531095605012.

\appendix

\section{GitHub link to code source} \label{sec:github}
The dynamic feature-based deep reinforcement learning (DF-DRL) model will be added to the DRLinFluids package. Please check in the following URL: \url{https://github.com/venturi123/DRLinFluids}. We invite all users to discuss further and ask for help directly on GitHub through the issue system, and we commit to helping develop a community around the DRLinFluids framework by providing in-depth documentation and help to new users. 

\section{Hyperparameters} \label{sec:hyperparameters}
Table \ref{tab:Hyperparameters_flow} and \ref{tab:Hyperparameters_DF-DRL} present the main numerical parameters of both the simulations and the learning algorithm.

\begin{table}
    \centering
    \caption{Configurations of flow simulation} \label{tab:Hyperparameters_flow}
        \begin{tabularx}{\textwidth}{ll}
        \toprule
        \textbf{Parameter} & \textbf{Value} \\
        \midrule
        Numerical time step (non-dimensional \dd{t})   & $ 5\times10^{-4}$ \\
        Maximum action amplitude (non-dimensional)   & 1.5 \\
        Action duration   & 25/44/46 \dd{t} for $Re$=100/500/1000 \\
        Number of action steps per episode & 100 \citep{wang2022drlinfluids} \\
        \bottomrule
        \end{tabularx}
\end{table}

\begin{table}
    \centering
    \caption{Hyper-parameters of the present DF-DRL model} \label{tab:Hyperparameters_DF-DRL}
        \begin{tabularx}{\textwidth}{ll}
        \toprule
        \textbf{Parameter} & \textbf{Value} \\
        \midrule
        Actor architecture   &  $512\times512$ (two fully connected layers) \\
        Critic architecture   & $512\times512$ (two fully connected layers) \\
        Actor leaning rate  &  $ 3\times10^{-4}$ \\
        Critic leaning rate  &  $ 2\times10^{-4}$ \\
        Discount factor  & 0.97 \\
        Alpha  & 0.2 \\
        Optimiser &  Adam \citep{kingma20153rd} \\
        Dynamic feature lifted states $\boldsymbol{S}$ & $\mathbb{R}^{30\times 2}$ (double vortex shedding periods in time dimension) \\
        CPU time per episode   & Up to 20 minutes \\
        Total CPU time of training stage   &  $\approx$ 5200 CPUh \\  
        \bottomrule
        \end{tabularx}
\end{table}

\bibliographystyle{jfm}
\bibliography{jfm}

\end{document}